\documentclass{article}
\usepackage{spconf,amsmath,graphicx}
\usepackage{url}

\usepackage[table,dvipsnames]{xcolor}
\usepackage{soul}
\usepackage{comment}
\usepackage{graphicx}
\usepackage{booktabs}
\usepackage{multirow}
\usepackage{subcaption}

\definecolor{Gray}{gray}{0.9}
\definecolor{lgray}{gray}{0.95}
\definecolor{LightCyan}{rgb}{0.92,0.92,1}
\definecolor{LightCyan}{rgb}{0.92,0.92,1}

\newcommand{\ie}{\emph{i.e.}~}
\newcommand{\eg}{\emph{e.g.,}~}
\newcommand{\etal}{\emph{et al.}~}

\def\ours{\emph{SimpleISP}~}

\definecolor{cvprblue}{rgb}{0.21,0.49,0.74}
\usepackage[pagebackref,breaklinks,colorlinks,citecolor=cvprblue]{hyperref}

\title{Simple Image Signal Processing using Global Context Guidance}
\name{Omar Elezabi, Marcos V. Conde, Radu Timofte}
\address{Computer Vision Lab, CAIDAS \& IFI, University of Würzburg}
\begin{document}
%
\maketitle
\begin{abstract}
 In modern smartphone cameras, the Image Signal Processor (ISP) is the core element that converts the RAW readings from the sensor into perceptually pleasant RGB images for the end users. The ISP is typically proprietary and handcrafted and consists of several blocks such as white balance, color correction, and tone mapping. Deep learning-based ISPs aim to transform RAW images into DSLR-like RGB images using deep neural networks. However, most learned ISPs are trained using patches (small regions) due to computational limitations. Such methods lack global context, which limits their efficacy on full-resolution images and harms their ability to capture global properties such as color constancy or illumination. First, we propose a novel module that can be integrated into any neural ISP to capture the global context information from the full RAW images. Second, we propose an efficient and simple neural ISP that utilizes our proposed module. Our model achieves state-of-the-art results on different benchmarks using diverse and real smartphone images.
\end{abstract}
\begin{keywords}
Image Processing, ISP, RAW, DSLR
\end{keywords}
%

\section{Introduction}
\label{sec:intro}

\begin{figure}[t]
    \centering
    \begin{tabular}{c}
         \includegraphics[trim={0cm 12cm 0cm 12cm}, clip, width=0.95\linewidth]{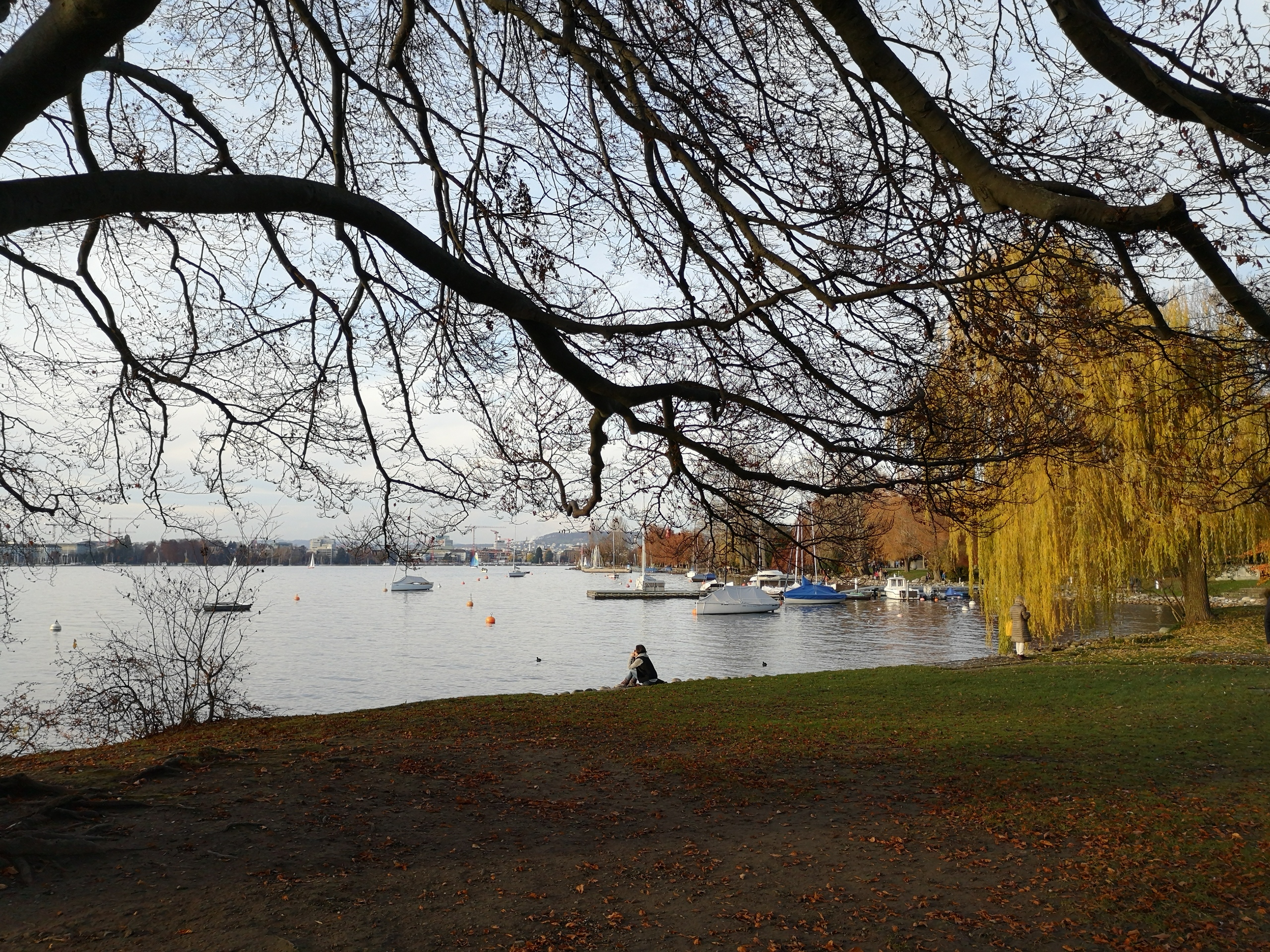} \\
         \includegraphics[trim={0cm 10cm 0cm 10cm}, clip, width=0.95\linewidth]{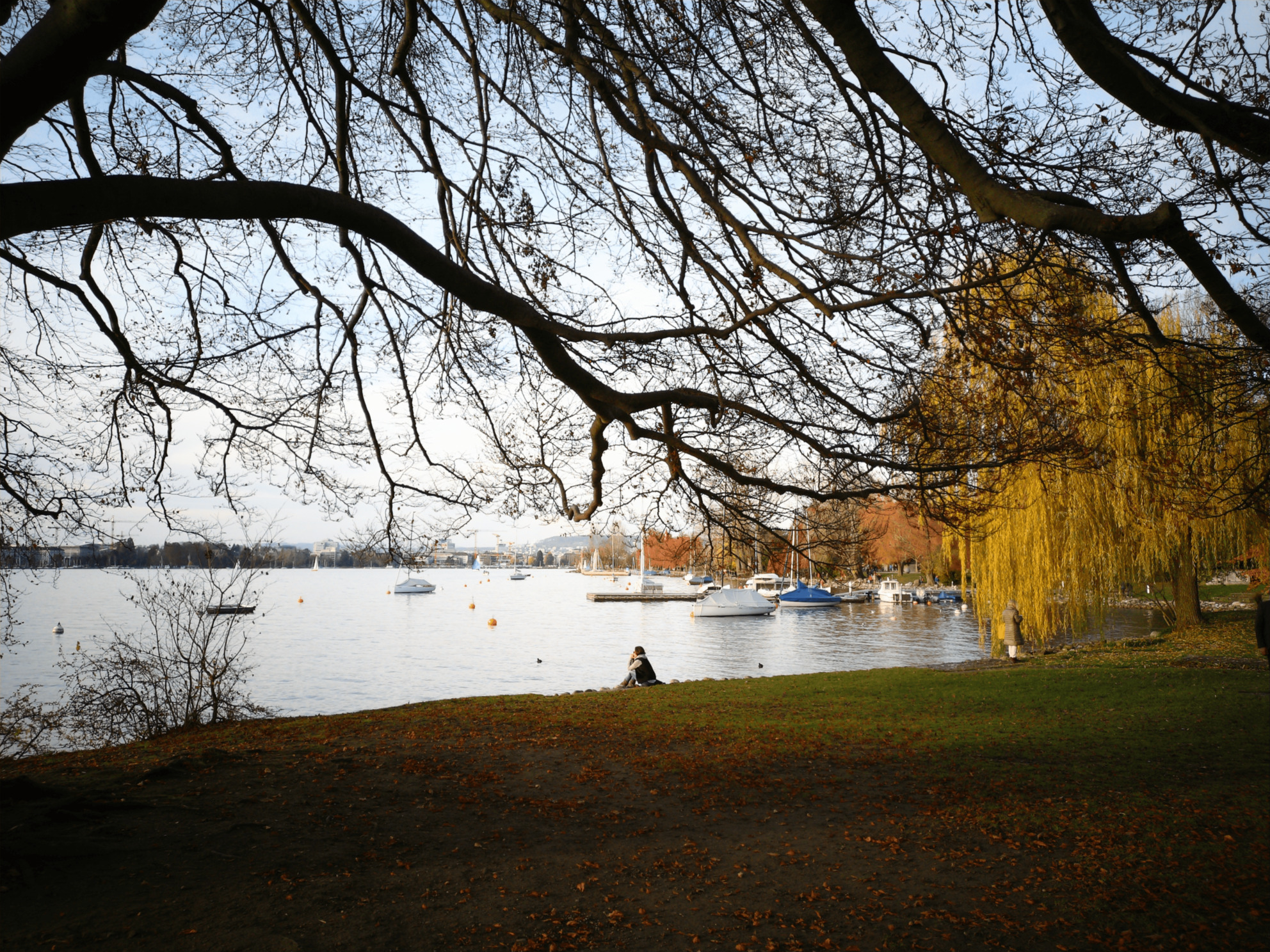} \\
    \end{tabular}
    \caption{(Up) Huawei P20 ISP and (Bot.) our \ours neural network. Both process the same Huawei P20 RAW image.}
    \label{fig:teaser}
\end{figure}

The Image Signal Processor (ISP) is a fundamental element in smartphone and DSLR imaging pipelines. The ISP converts the sensor RAW image (linear \emph{w.r.t} the scene radiance) into high-quality RGB images. The ISP is typically a model-based pipeline consisting of several blocks such as denoising, demosaicing, color constancy, and color space transformations~\cite{delbracio2021mobile, conde2022model}. Nowadays, many of these building blocks are implemented as neural networks.

Deep learning-based ISPs~\cite{Ignatov2020pynet, liang2019cameranet, ignatov2022learned, Liu2022, ignatov2022microisp} aim to learn the RAW-to-RGB transformation as an end-to-end differentiable function \ie a deep neural network. Aiming at surpassing current smartphone ISPs, some approaches map smartphone RAW images to RGB images captured with DSLR cameras~\cite{Ignatov2020pynet, ignatov2022microisp, shekhar2022transform}. Such recent approaches achieved notable results thanks to new deep learning models and datasets~\cite{Ignatov2020pynet, shekhar2022transform, ignatov2021learned}. 

\begin{figure}[t]
     \centering
     \setlength\tabcolsep{1pt}
     \begin{tabular}{c c}
          \includegraphics[width=0.49\linewidth]{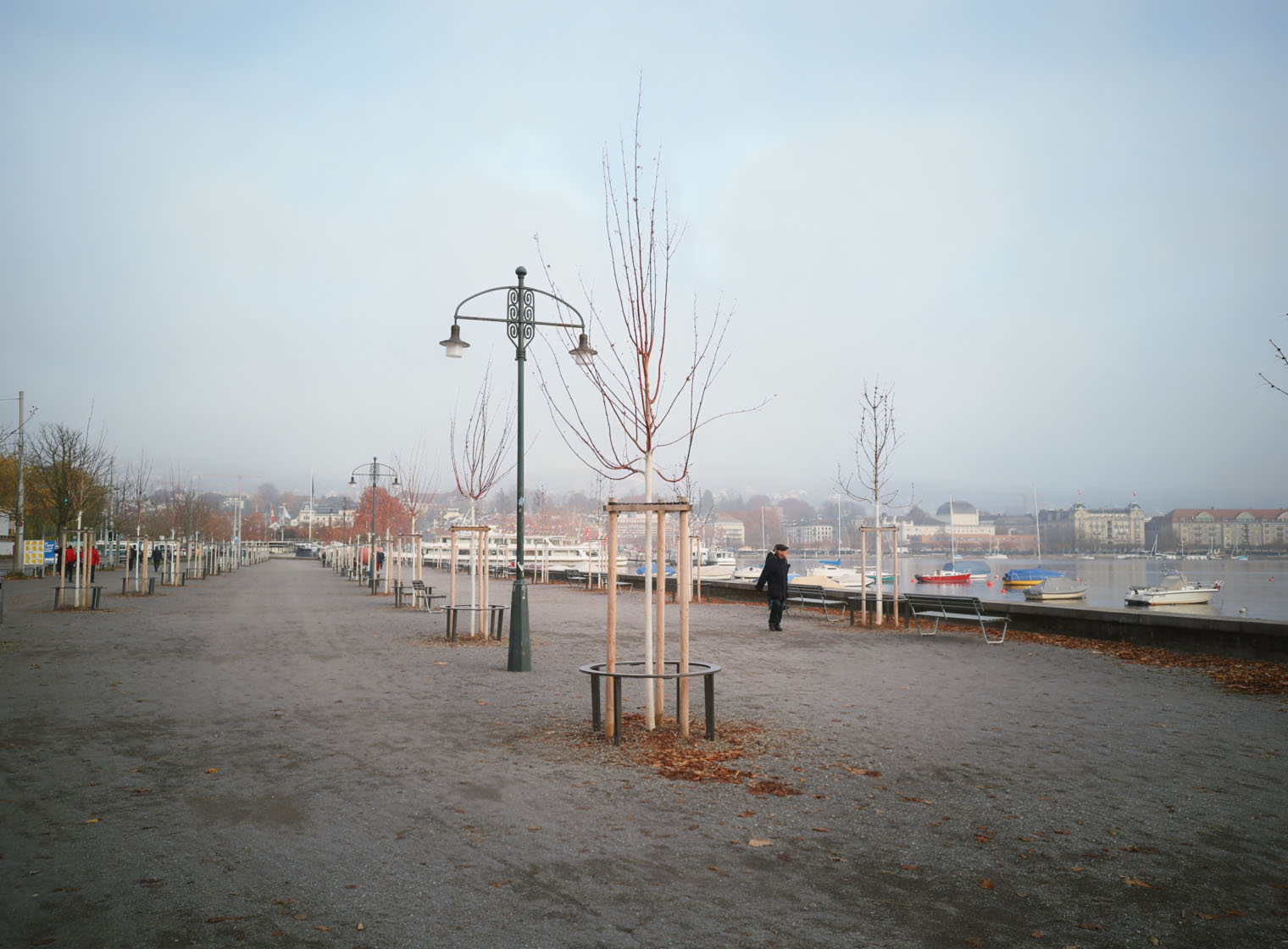} &
          \includegraphics[width=0.49\linewidth]{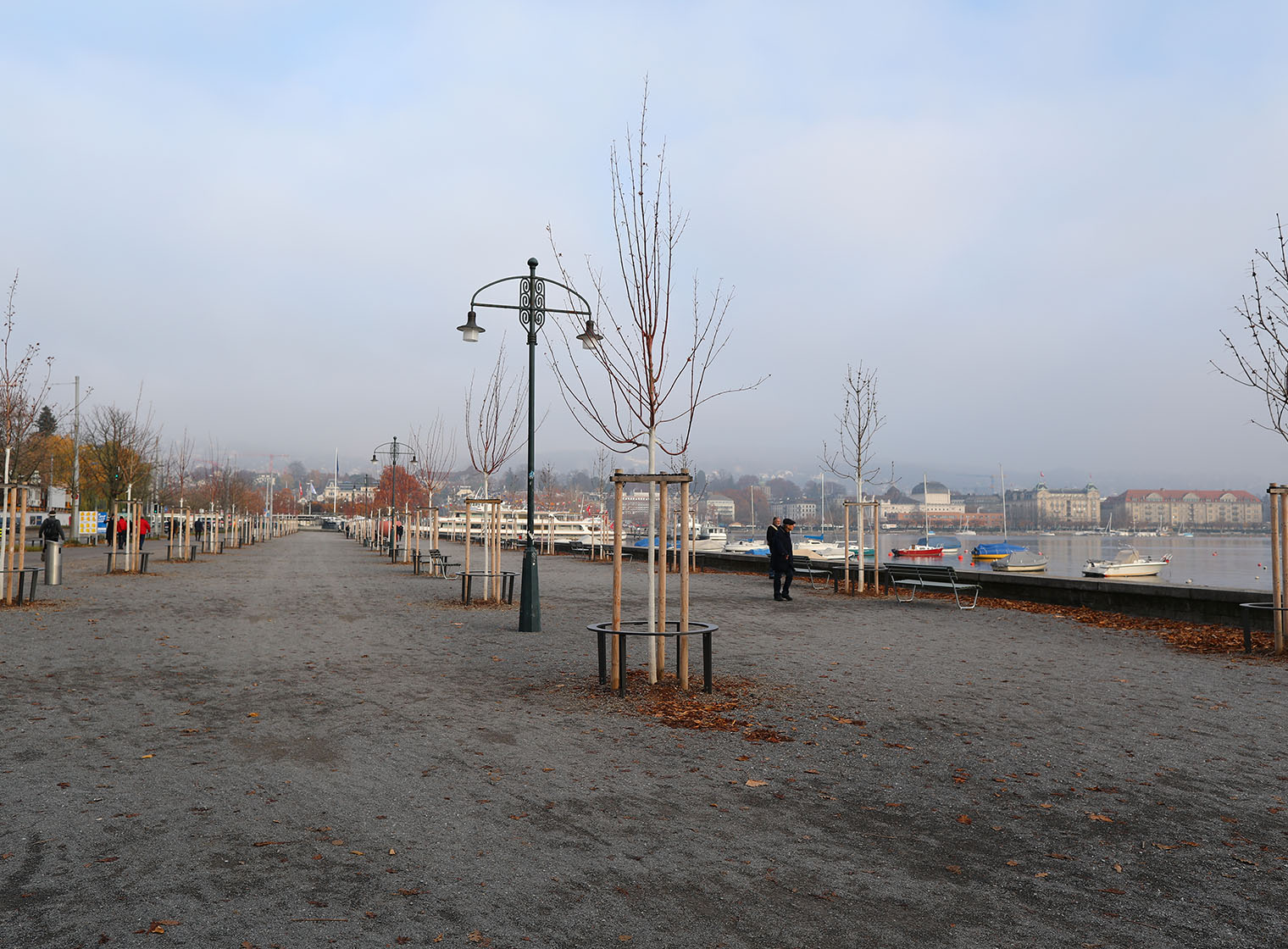} \\
          Neural ISP~\cite{Ignatov2020pynet} & DSLR Target \\
     \end{tabular}
     \caption{Example of the limitations of learned ISPs trained using patches. We can appreciate inconsistent vignetting, non-uniform illumination and colors (see the center and sky).}
     \label{fig:pynet}
\end{figure}

However, we detect a \emph{limitation} in most learned ISPs. Training complex deep learning ISP methods on high-resolution (HR) images (\eg 12MP) is very time consuming and computationally expensive. Even high-performance GPUs struggle to allocate the required memory. For this reason, most methods are trained -and evaluated- using image patches (small regions \eg 256px). Yet, such lack of global context -the whole image- limits their performance on HR images, and harms their ability to capture global properties such as global illumination~\cite{cheng2014illuminant}. For instance, such methods tend to produce inconsistent colors and illumination across the image. We provide an example in Figures~\ref{fig:teaser} and~\ref{fig:pynet}.

Therefore, we assume that patch-based training is limited \ie the model lacks global information, and we aim to solve this problem. Our new training method incorporates global context (attending to the whole image) into the patch-based training and improves the performance of different ISP models consistently. To prove this, we created two new datasets based on public sources to ensure reproducibility and fair comparisons. We also tested our method on a RAW image super-resolution dataset~\cite{zhang2019zoom} with task joint RGB reconstruction and super-resolution. Additionally, by utilizing this method, we propose a new simpler, and more efficient ISP than current models while achieving state-of-the-art performance.

\section{Related Work}
\label{sec:rel_work}

As in previous works, we define the ISP transformation as:

\begin{equation}
    x = \mathcal{F}(y) \quad \text{\small where}~~ x \in R^{H\times W\times 3} ~,~ y \in R^{\frac{H}{2}\times \frac{W}{2}\times 4}
\label{eq:isp}
\end{equation}

where $x$ is an RGB image, and $y$ is a RAW image that follows a 4-channel Bayer pattern (RGGB). The ISP $\mathcal{F}$ is considered a composition of non-linear functions. Following~\cite{Ignatov2020pynet, ignatov2022learned, shekhar2022transform, zhang2021learning} we focus on smartphone data.

Deep learning models are commonly used to model $\mathcal{F}$ \ie the ISP transformation from RAW to RGB. A single neural network is mostly used to model the full ISP~\cite{Ignatov2020pynet, zhang2021learning, xing2021invertible}. Some methods use multi-stage or multi-network models with neural networks that tackle different categories of tasks separately~\cite{shekhar2022transform, liang2019cameranet, Dai2020}, or multiple neural networks with specific tasks combined together~\cite{conde2022model, Liu2022}.

These methods demonstrate the capability of neural networks to replicate complex non-linear ISP transformations. However, there are certain limitations associated with these deep learning approaches: (i) The requirement for extensive datasets to achieve effective generalization is a challenging endeavor given the scarcity of paired RAW-RGB smartphone data~\cite{Ignatov2020pynet,ignatov2021learned}. (ii) The complexity of these models makes them unsuitable for deployment on mobile devices~\cite{ignatov2022learned, ignatov2021learned}. (iii) Despite advancements, the quality of the resulting RGB images has yet to reach the quality of DSLR photos. (iv) Such deep models learn $\mathcal{F}$ as black boxes, lacking interpretability and explicit control of the transformation.

Ignatov~\etal introduced the Zurich RAW-to-RGB (ZRR) Dataset~\cite{Ignatov2020pynet} and the Fujifilm Dataset~\cite{ignatov2021learned} that consist of RAW images captured with smartphone sensors (\eg Huawei P20 phone) and RGB images captured using professional DSLR cameras as the ground-truth (\eg Canon D5 Mark IV). Such datasets are used in multiple ``Learned ISP Challenges"~\cite{ignatov2022learned, ignatov2021learned} focused on designing neural ISPs suitable for mobile devices. 

All available ISP methods utilize image patches for training which eliminates the use of global information from the full scene. Some methods utilize global operations such as channel attention~\cite{shekhar2022transform, woo2018cbam, Hsyu2021} which includes some global information in the network, however, these methods are still limited to a small region of the image (the patch) since this is the only information available during training. 

\section{Proposed method}
\label{sec:ours}
\begin{figure}[t]
\centering
\includegraphics[width=\linewidth]{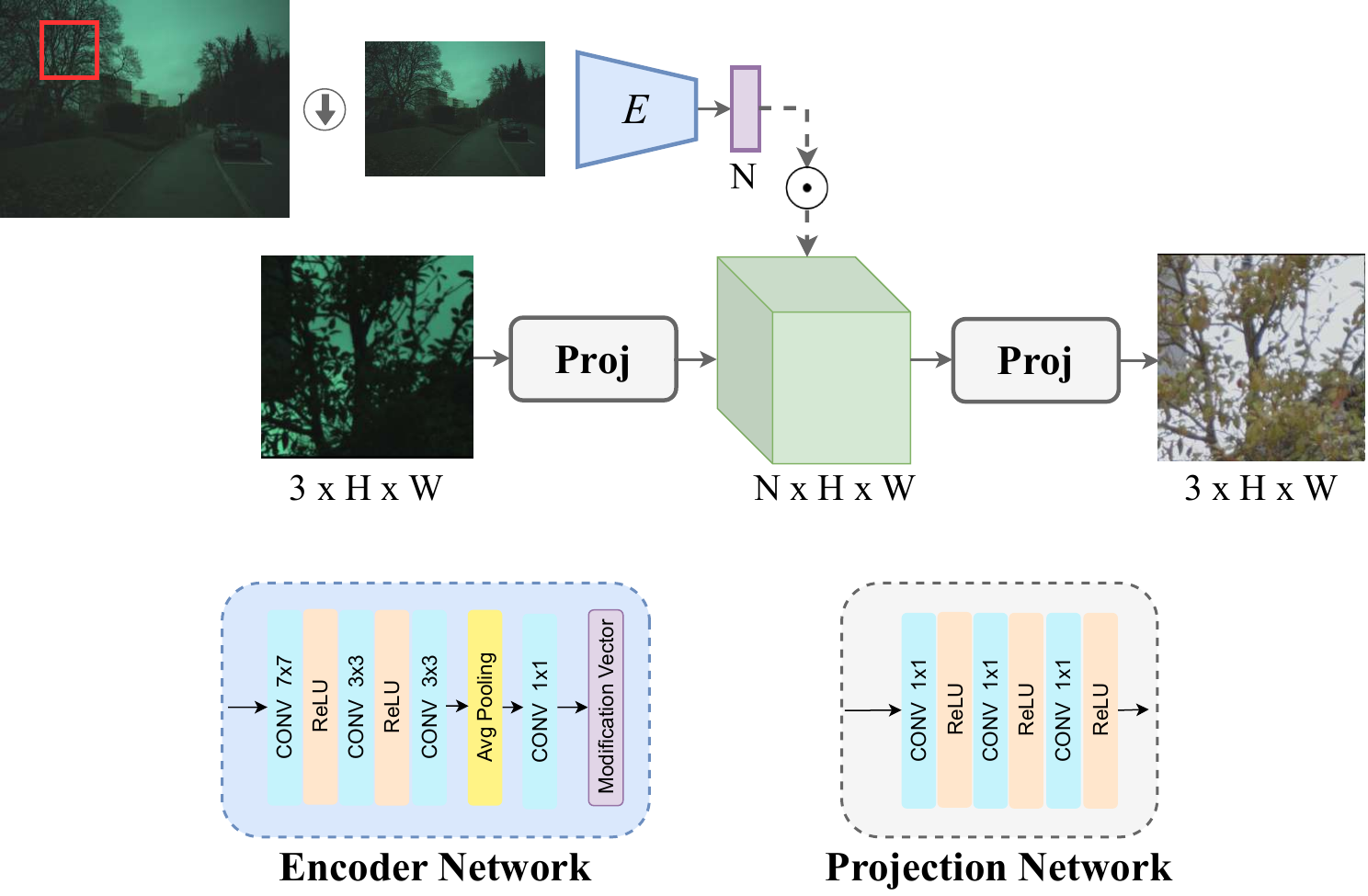}
\caption{This figure shows the architecture of our proposed CMod (Top). $E$ is the encoder network that takes the resized full RAW image as input and produces the modification vector. \textsc{Proj} is the projection network that projects the input to the modification space, and then back to the RGB space. The \emph{global guidance} is applied by a channel-wise multiplication $\odot$ between the projected image and the modification vector. 
}
\label{fig:cmod_comps}
\end{figure}

\begin{figure*}[!ht]
\centering
\includegraphics[width=\textwidth]{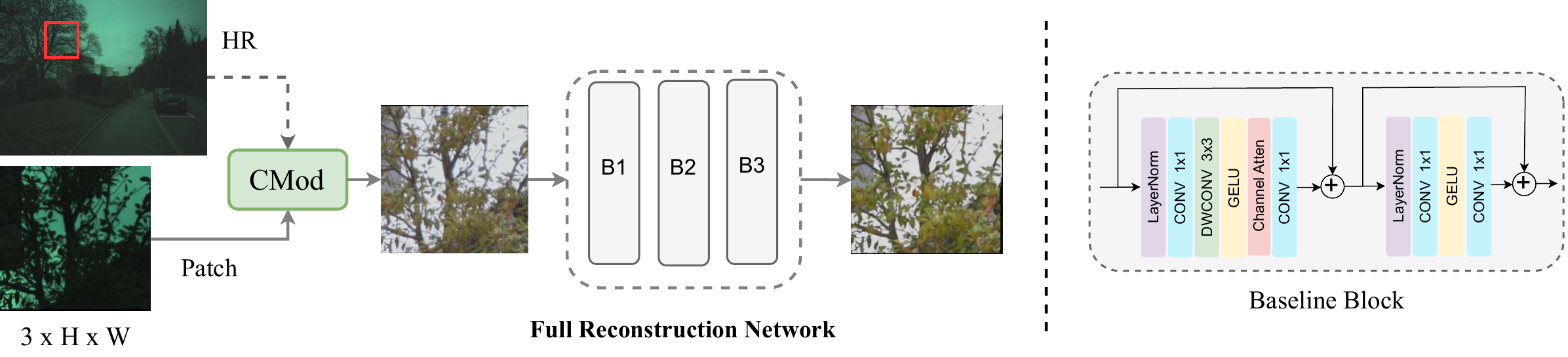}
\caption{This figure shows the full pipeline of our proposed ISP model, \ours. First, we feed the RAW image into the CMod module for color reproduction. Then the output of CMod is processed by the full reconstruction network to produce the final RGB output. This network is conformed by three blocks B$_i$ inspired by Chen~\etal ~\cite{chen2022simple}. We illustrate these building blocks --the Baseline block-- on the right side. We build \ours using three baseline blocks.
} 
\label{fig:cmod-dia}
\end{figure*}

We propose a new training approach to include global context information in patch-based training. We achieve that by adding our proposed Color Module (CMod) that can be integrated in any ISP Network. We illustrate the overall neural ISP in Figure~\ref{fig:cmod-dia}. We explain the novel CMod module in Section~\ref{sec:global_training}, and we illustrate it in Figure~\ref{fig:cmod_comps}.

In Section~\ref{sec:simpleis} we introduce our novel neural ISP that serves as an efficient and simple baseline for this task.

\subsection{Color Module}
\label{sec:global_training}
We created a \emph{Color module (CMod)} that is added before the RGB reconstruction branch for the purpose of color reconstruction. We perform first the color reconstruction task, as it is mainly a global modification task that depends on the global information in the image. 
Such global modification mainly consists of white balancing and color correction. These tasks are usually pixel-wise modifications. To follow this process we use a pixel-wise network consisting of $1\times1$ convolutional layers to ensure only pixel modifications.

Our module consists of 3 parts as seen in Fig. \ref{fig:cmod_comps}. The first part projects the input RAW image into the modification space that embeds each pixel into a k-dimensional vector by $1\times1$ convolution layers. The second part is the modification vector that is created by an encoding network. That network takes the guidance image and encodes it into a k-dimensional vector. That modification vector will be used to apply the color modification to the input image projected into the modification space by direct multiplication. The modification vector is created by an encoder network ($E$) which uses large kernel convolutions and average pooling. After the multiplication operation, we project the image again to the RGB space using another projection network.

The projection process to higher feature space allows us to apply different color modification operations like white balance and color correction with just a simple vector multiplication. For our modification space size, we chose 64 as the size of the vectors. 

We use the resized full RAW image as the guidance image. To resize the full raw image we separate the different color bands into 4 channels, and resize each band separately. This avoids mixing information of different bands and destroying the raw pattern. 

This pipeline allows us to utilize the global information from the whole image with a small computation cost by encoding the full image into just a modification vector. The next equations describe the operations in the CMod block:

\begin{equation}
    X_m = P_{{raw} \rightarrow M}~(X)
\end{equation}
\begin{equation}
    mv = E(G)
\end{equation}
\begin{equation}
    X_{mv} = X_m \odot mv
\end{equation}
\begin{equation}
    y^c = P_{M \rightarrow {rgb}}~(X_{mv})
\end{equation}
Where $P$ is the projection network, $E$ is the encoding network, $X$ is the RAW patch image, $G$ is the guidance image, $mv$ is the modification vector, $X_m$ is the input RAW patch in the modification space and $X_{mv}$ after applying the modification vector. Finally, $y^c$ is the output of the CMod (see Fig.~\ref{fig:cmod_comps}).

The Final ISP Network starts with CMod which reproduces a colored image without any texture or content modifications just pixel-wise color adjustment. This colored image is the input to the Reconstruction Network which reconstructs the final image. The input for our model is a simple demosaiced version using bi-linear interpolation of the input RAW patch. We use a demosaiced image to ensure the same size throughout the network to be able to compare the output of the CMod Network with the ground-truth. 

For the final inference on the full image, the input will be the resized full RAW image as a guide image and the full RAW image in full resolution for the ISP Network.

\subsection{Designing an ISP with Global Context}
\label{sec:simpleis}

We utilize our CMod to create a very efficient ISP network while maintaining \emph{state-of-the-art} performance. We split the ISP network into 2 different modules, CMod for global modifications, and the main branch for the full RGB reconstruction. These 2 modules are connected and trained in an end-to-end fashion. Splitting global and local modifications allows us to use a very efficient network for the full reconstruction stage. This differs from previous methods that require complex neural networks for learning both the global and local transformations of the ISP. 
To build our efficient ISP model we utilize the \emph{baseline block} from Chen~\etal ~\cite{chen2022simple}. The baseline block is used to create a general-purpose image restoration model, as this block is designed to perform efficient local operations and learn spatial and channel interactions. This block is extremely efficient and has proven to be very powerful in restoration tasks.

The \emph{baseline block} consists of an inverted residual block ~\cite{sandler2018mobilenetv2} enhanced with Layer Normalization~\cite{ba2016layer} --for training stabilization--, and channel attention~\cite{woo2018cbam}. We also use non-linear GELU activation as previous works~\cite{chen2022simple}.

We illustrate the structure of the final network in Fig. \ref{fig:cmod-dia}

\section{Experimental Results}
\label{sec:results}

\subsection{Datasets}
\label{sec:datasets}

\textbf{Neural ISP}: Our approach aims at utilizing the full-resolution RAW images for better global properties and more accurate reconstruction. However, most datasets only provide image patches without the full images. We built upon the Zurich RAW-to-RGB (ZRR) dataset~\cite{Ignatov2020pynet} and the ISPIW dataset~\cite{shekhar2022transform} to create datasets with both patches and full-resolution RAW images.

For these datasets, we find the following \textbf{limitations}: (i) the original ZRR dataset~\cite{Ignatov2020pynet} does not provide full-resolution RAW images, only the test set includes the full RAW images and for the rest of the dataset only image patches are available. (ii) the  ISPIW dataset~\cite{shekhar2022transform} does not provide the code for the pre-processing of patches they provided or the reference train-test split. For these reasons, reproducing results on the original dataset benchmarks is not possible. (iii) Moreover, as previously discussed, the RAW-RGB pairs are not perfectly aligned.

Due to such limitations, we create our versions of these datasets, such that the results are fully reproducible using patches or high-resolution RAW images. 

We create our own training and testing splits from the available full RAW and DSLR image pairs. To create the image patches for both datasets we followed the same process as~\cite{Ignatov2020pynet}. We first align the image pairs using SIFT~\cite{lowe2004distinctive} and RANSAC~\cite{fischler1981random}. Then split the images into RAW-RGB patch pairs of size $448\times448$. 
In order to avoid the pairs with extreme misalignment, we only considered the pairs with cross-correlation $>0.5$. For each patch, we have the associated full-resolution image.

\vspace{2mm}
\textbf{ZRR Small}: The dataset consists of RAW images from the Huawei P20 smartphone and RGB reference images from the Canon 5D Mark IV camera. We utilize the 148 full-resolution images from ZRR dataset \cite{Ignatov2020pynet} with a 90/10 split to create our dataset. After filtering, this results in 5599 patch pairs for training and 596 for testing. 

\vspace{2mm}
\textbf{ISPIW}: The original dataset~\cite{shekhar2022transform} consists of RAW images from the Huawei Mate 30 Pro smartphone and RGB reference images from the Canon 5D Mark IV camera. 
We utilize the available 192 full-resolution images from this dataset with a 90/10 split to create our version. This results in 5152 patch pairs for training and 572 for testing.

\vspace{2mm}
\hspace{-6mm} \textbf{Raw Super Resolution}:
For a more comprehensive test, we also evaluated our module on additional tasks. We tested our module joint RGB Reconstruction and Super Resolution. We choose this task to show the importance of global context on any reconstruction task and the improvements of including global context even for local modifications. The dataset consists of low-resolution RAW images as input and high-resolution RGB images as output. 

\textbf{SR-RAW \cite{zhang2019zoom}}: the dataset consists of 500 scenes of images. for each scene different focal lenses were captured (24, 35, 50, 70, 100, 150, and 240 mm) to construct the super resolution data. The 24/100, 35/150, and 50/240 pairs were used to form a 4× super-resolution dataset. We followed the same process as \cite{zhang2021learning} by using 400 scenes for training, and 50 for validation. Then the performance was reported on 35/150 mm image pairs of the remaining 50 sences. The dataset included all the required elements for our experiment, including the full raw input image, so no additional processing was required.

\subsection{Experimental Setup}

In our experiments, we use our created datasets to benchmark our proposed model, as well as other state-of-the-art methods.

For a fair comparison, we re-train all the models using our datasets and a common setup. We also consider the specific training requirements of each model. We use their official implementations when available.

\vspace{2mm}
\textbf{Data Misalignment Issue}: For the creation of the dataset we choose a low cross-correlation threshold to include more patches in the dataset. This results in samples with low pixel alignment. For more accurate training and evaluation, we utilize the miss-alignment training process from Zhang \etal~\cite{zhang2021learning} to minimize the issues from training with miss-aligned patches and focus on the models' reconstruction performance. We also utilize the same optical flow module to densely align the test patches, leading to a more accurate benchmark. We use this setup for all tested models for a fair comparison.

\vspace{2mm}
\textbf{Implementation Details}: The only difference in the training setup between the models is the representation of the input. For the models that contain CMod, the input is a simply demosaiced version of the RAW image. The other models or variants use as input the 4-channel RAW. We train our models for 200 epochs with Adam~\cite{kingma2014adam} optimizer with a learning rate of $1e^{-4}$, and a learning rate decay of $0.5$ every 40 epochs. The models are trained using a batch size of 4 on NVIDIA GeForce RTX 4090 24 GB GPU.

\textbf{Loss Functions}: For CMod we optimize the output of the module using only the color loss, since the focus of this module is color reproduction. We use the color difference between the output from the module and the reference image. We use CDNet ~\cite{wang2023measuring} pre-trained model as our color loss. For the final output, we use a combination of MSE loss, VGG-based ~\cite{johnson2016perceptual} perceptual loss, SSIM loss, and color loss.

\subsection{Ablation studies}

\begin{table}[t]
    \centering
    \resizebox{\linewidth}{!}{
    \begin{tabular}{l c c c c}
        \toprule
        Method & PSNR $\uparrow$ & SSIM $\uparrow$ & LPIPS $\downarrow$ & E00 $\downarrow$\\
        \midrule
        Baseline (LiteISP)~\cite{zhang2021learning} & 22.18 & 0.8305 & 0.162 & 11.610\\
        Baseline With CDNet  & 22.37 & 0.8303 & 0.166 & 11.296\\
        Ours with RAW patch guide  & 22.54 & 0.8441 & 0.146 & 11.049\\
        \rowcolor{lgray} Ours with full RAW image guide & \textbf{24.79} & \textbf{0.8593} & \textbf{0.135} & \textbf{9.486}\\
        \bottomrule
    \end{tabular}
    }
    \caption{Ablation study of our modifications to the baseline LiteISP~\cite{zhang2021learning}. This experiment was conducted on \emph{\textbf{ZRR Dataset}~\cite{Ignatov2020pynet}} \emph{(Huawei P20)}.}
    \label{tab:Cmod}
\end{table}

\begin{figure}[t]
    \centering
    \setlength{\tabcolsep}{1pt}
    \begin{tabular}{ccccc}
         \includegraphics[width=0.19\linewidth]{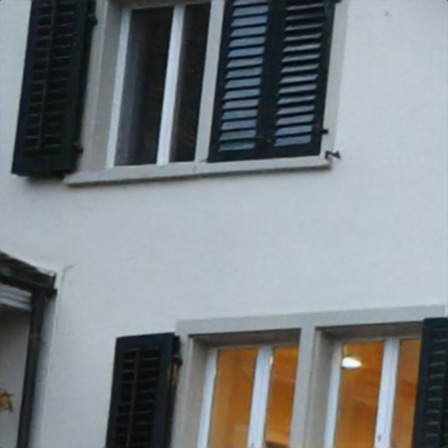} &
         \includegraphics[width=0.19\linewidth]{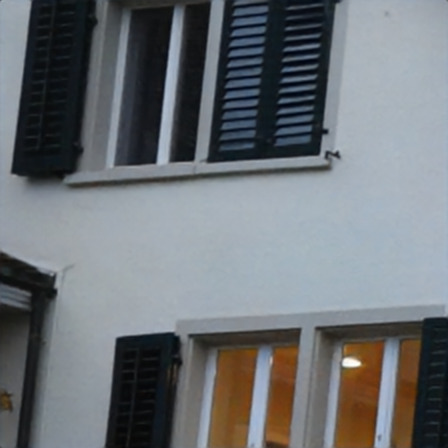} &
         \includegraphics[width=0.19\linewidth]{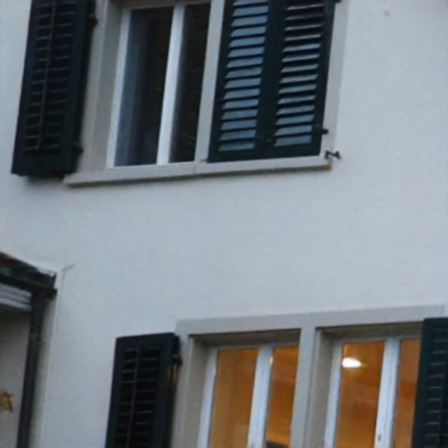} &
         \includegraphics[width=0.19\linewidth]{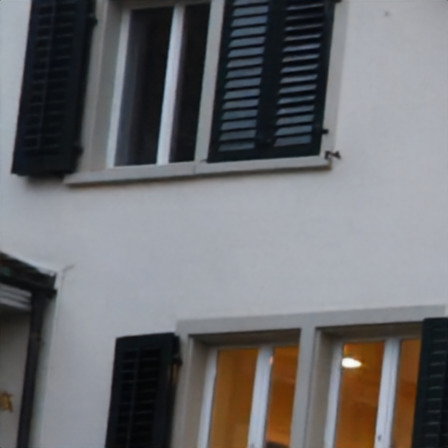} &
         \includegraphics[width=0.19\linewidth]{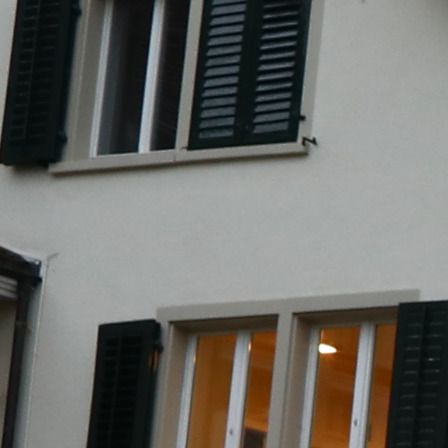} \\
         \includegraphics[width=0.19\linewidth]{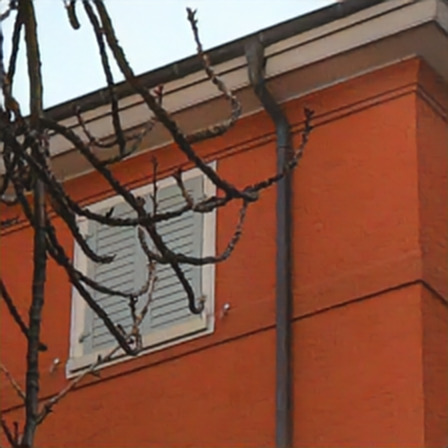} &
         \includegraphics[width=0.19\linewidth]{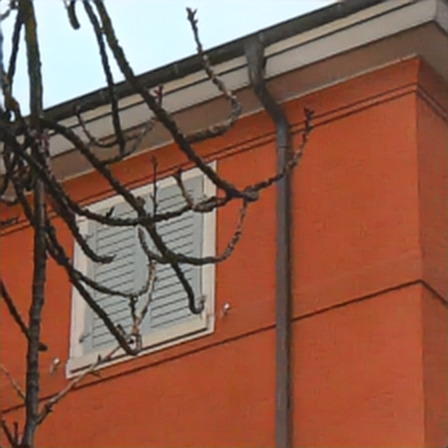} &
         \includegraphics[width=0.19\linewidth]{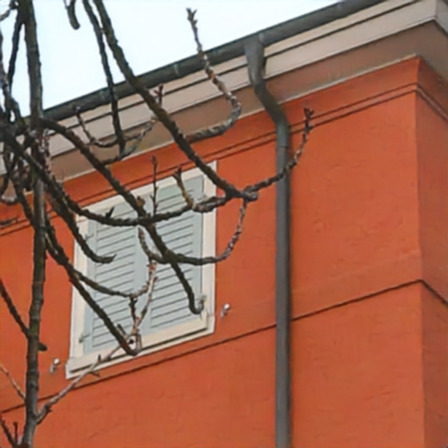} &
         \includegraphics[width=0.19\linewidth]{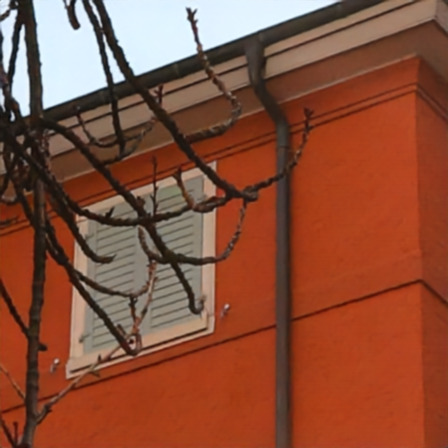} &
         \includegraphics[width=0.19\linewidth]{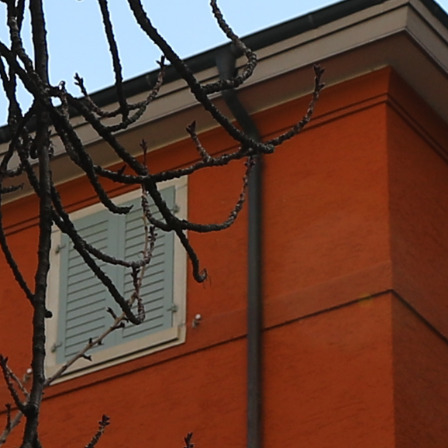} \\
         \includegraphics[width=0.19\linewidth]{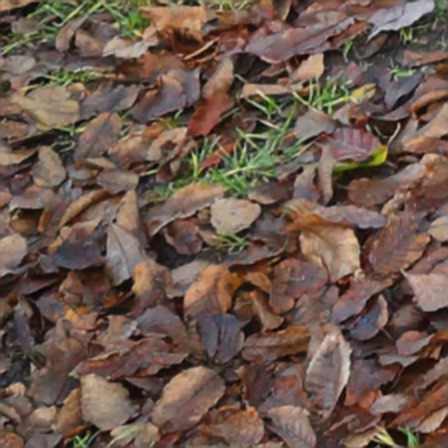} &
         \includegraphics[width=0.19\linewidth]{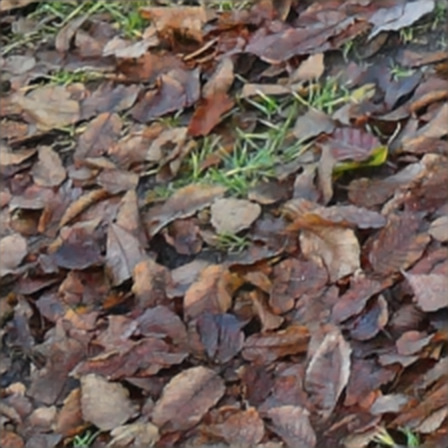} &
         \includegraphics[width=0.19\linewidth]{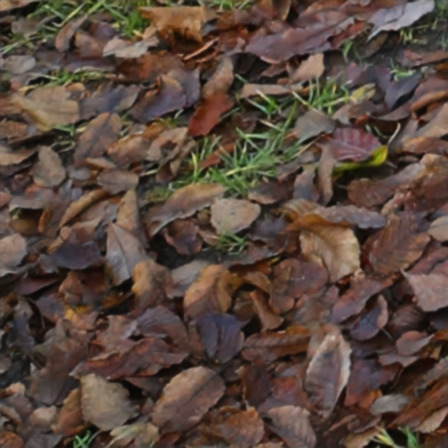} &
         \includegraphics[width=0.19\linewidth]{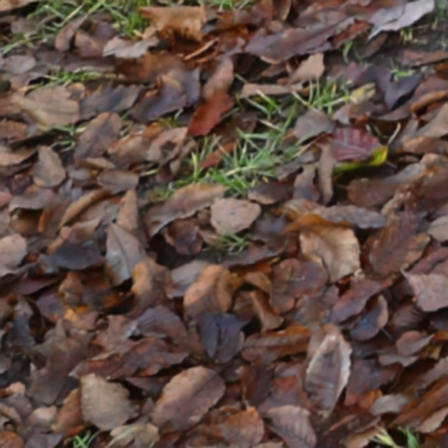} &
         \includegraphics[width=0.19\linewidth]{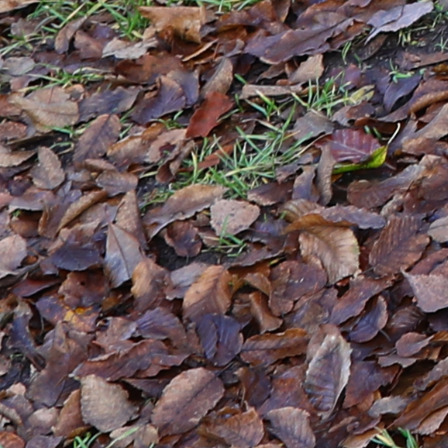} \\
         \small{BaseLine} & \small{+ CDNet} & \small{+ P. Guide} & \small{+ FI. Guide} & \small{DSLR GT} \\
    \end{tabular}
    
\caption{Visual ablation study on the \emph{effect of our global context module}. We use as baseline LiteISP~\cite{zhang2021learning}. We show the results using the Patch (P.) guidance, and the Full-resolution Image (FI.) guidance. Our global guidance improves notably the results.}
\label{fig:cmod_output}
\end{figure}

\subsubsection{CMod with Neural ISP}

As we show in Tab.~\ref{tab:Cmod}, we study the performance of the model after each proposed modification. 

First, we evaluate the performance of the baseline model (LiteISP~\cite{zhang2021learning}) on our dataset.

Second, we study the addition of a specific color loss -CDNet~\cite{wang2023measuring}- to improve the color reproduction of the model. 

Third, we study the benefit of learning global operations such as white balance and color correction separately with a specific module (CMod). Thus, we add our CMod to the baseline to create the guidance vector using the RAW image patch itself as input.

Finally, we study the importance of having a full view of the scene for global operations. Thus, we use the full-resolution RAW image as the guidance image. 

In Tab.~\ref{tab:Cmod} we summarize this ablation study. We can observe the benefits of adding the CD loss (CDNet)~\cite{wang2023measuring}, as it adds more explicit weight to the optimization of color matching, at no cost.
Then we can appreciate how CMod guidance helps to improve the performance. In particular, using the full-resolution images improved the performance of our model drastically ($+2.2$ dB) \emph{w.r.t} the baseline. We can conclude that training using only image patches is very limiting, especially with the global tasks within the ISP. Therefore, our best models include CMod using the full image, and additional color loss. In Tables \ref{tab:results-p20} and \ref{tab:results-ispw}, we can see the benefits of adding CMod to different neural ISPs on different datasets. 

For the \emph{quantitative results}, in Fig. \ref{fig:cmod_output} we compare the output of the baseline model and its different modification stages (same as in Tab. \ref{tab:Cmod}). We can appreciate how each technique improves progressively the performance in terms of color accuracy and texture. Our final model using the full-resolution image as guidance and color loss is the best in all aspects. 


\begin{table}[t]
    \centering
    \resizebox{\linewidth}{!}{
    \begin{tabular}{l c c c c c}
        \toprule
        Method & \# Params  & MACs  & Time (Patch) & Time (HR) & PSNR \\
        & (M) & (G) & (ms) & (ms) & (dB)\\
        \midrule
        LiteISP~$\dagger$              & 8.9 & 261.77 & 0.036  & fail & 24.79 \\
        MicroISP~$\dagger$             & 0.045 & 5.14 & 0.006 & 0.026 & 22.04 \\
        \rowcolor{lgray} \ours     & 0.064  & 8.42 & 0.011 & 0.035 & 24.18 \\
        \bottomrule
    \end{tabular}
    }
    \caption{Efficiency study. The models with $\dagger$ were trained using our CMod for global context. We compare the models' complexity in terms of parameters, operations (MACs), and inference time on both patch and full-resolution images. We calculate MACs using $448\times448$ input patches. We measured inference time in ms on the NVIDIA RTX 3090 24GB GPU.}
    \label{tab:cmod-time}
\end{table}

\vspace{2mm}
\textbf{Efficiency Study} In Table~\ref{tab:cmod-time} we compare the efficiency of our \ours model to the improved LiteISP~$\dagger$. We can see that our model is far more efficient with $20\times$ fewer parameters and operations. 


\subsubsection{CMod for Raw Super Resolution}

We tested the addition of our module with full-resolution guidance on the task of RAW super-resolution~\cite{conde2024bsraw}. For the base model, we choose the Zhang \etal~\cite{zhang2021learning} super-resolution model which is based on the SRResNet~\cite{ledig2017photo} model.

Tab. \ref{tab:sraw} shows the performance of the base model without and with our CMod. As we notice from the results our CMod largely improves the performance of the model with an increase of 0.2 dB. Also, the results show a huge improvement in image structure and quality as we notice from the SSIM and LPIPS metrics.

The qualitative results in Fig. \ref{fig:sraw-results} validate these results. As we notice in the baseline output, the reconstructed images are overall darker with worse color rendering and illumination processing. On the contrary, the baseline model with the addition of CMod produces improved results with better and more natural colors and improved illumination handling. This shows the importance of global information for any reconstruction model. We can also appreciate the improvement in the local features processing with improved texture, details, and overall quality.

\begin{table*}[t]
    \centering
    \begin{subtable}[t]{0.49\textwidth}
    \centering
    \begin{tabular}{l c c c}
         \toprule
         Method & PSNR~$\uparrow$ & SSIM~$\uparrow$ & LPIPS~$\downarrow$ \\
         \midrule
         LiteISP~\cite{zhang2021learning}                  & 22.18 & 0.8305 & 0.162 \\
         MicroISP~\cite{ignatov2022microisp}                 & 20.30 & 0.7826 & 0.264 \\
         AWNet~\cite{Dai2020}                   & 21.80 & 0.8205 & 0.184 \\
         MW-ISPNet~\cite{ignatov2020aim}                & 21.88 & 0.8234 & 0.178 \\
         \midrule
         LiteISP~$\dagger$        & \textbf{24.79} & \textbf{0.8593} & \textbf{0.135} \\
         MicroISP~$\dagger$       & 22.04 & 0.8145 & 0.255 \\
         \rowcolor{lgray}\ours~(\emph{Ours})  & \underline{24.18} & \underline{0.8432} & \underline{0.173} \\
         \bottomrule
    \end{tabular}
    \caption{\emph{\textbf{ZRR Dataset}~\cite{Ignatov2020pynet} (Huawei P20)}.}
    \label{tab:results-p20}
    \end{subtable}
    \hfill
    \begin{subtable}[t]{0.49\textwidth}
    \centering
    \begin{tabular}{l c c c}
         \toprule
         Method & PSNR~$\uparrow$ & SSIM~$\uparrow$ & LPIPS~$\downarrow$ \\
         \midrule
         LiteISP~\cite{zhang2021learning}                  & 22.14 & 0.8146 & 0.207 \\
         MicroISP~\cite{ignatov2022microisp}                 & 20.70 & 0.7735 & 0.307 \\
         AWNet~\cite{Dai2020}                    & 21.62 & 0.8071 & 0.212 \\
         MW-ISPNet~\cite{ignatov2020aim}                & 21.90 & 0.8102 & 0.217 \\
         \midrule
         LiteISP~$\dagger$        & \textbf{24.04} & \textbf{0.8409} & \textbf{0.161} \\
         MicroISP~$\dagger$       & 22.48 & 0.8105 & 0.268 \\
         \rowcolor{lgray}\ours~(\emph{Ours})  & \underline{23.67} & \underline{0.8247} & \underline{0.210} \\
         \bottomrule
    \end{tabular}
    \caption{\emph{\textbf{ISPIW Dataset}~\cite{shekhar2022transform} (Huawei Mate 30 Pro)}}
    \label{tab:results-ispw}
    \end{subtable}

\caption{RAW-RGB mapping results using smartphone datasets. The metrics are calculated between the resultant RGB images from each neural ISP, and the reference DSLR images. The methods with $\dagger$ were trained with the addition of our global context guidance (CMod). Note that our \ours is more efficient than LiteISP~\cite{zhang2021learning}. We highlight the \textbf{best} and \underline{second best}}
\label{tab:benchmarks}
\end{table*}

\begin{table}[t]
    \centering
    \resizebox{0.9\linewidth}{!}{
    \begin{tabular}{l c c c c}
        \toprule
        Method & PSNR $\uparrow$ & SSIM $\uparrow$ & LPIPS $\downarrow$\\
        \midrule
        Zhang \etal~\cite{zhang2021learning} & 21.18 & 0.6828 & 0.365\\
        Ours  & 21.36 & 0.7509 & 0.336 \\
        \bottomrule
    \end{tabular}
    }
    \caption{RAW Super Resolution results using SR-Raw Dataset \cite{zhang2019zoom}. The metrics are calculated between the resultant RGB images from each network, and the reference GT images. Zhang \etal~\cite{zhang2021learning} is the baseline model. Ours is the baseline model with the addition of CMod.}
    \label{tab:sraw}
\end{table}

\subsection{Benchmarks}

\textbf{State-of-the-art comparison}: In Tab.~\ref{tab:benchmarks} we compare our \ours model with other \emph{state-of-the-art} neural ISP methods. We also validate our results using the 2 proposed benchmark datasets, \emph{ZRR Small} (Tab.~\ref{tab:results-p20}) and \emph{ISPIW} (Tab.~\ref{tab:results-ispw}). As we see in Tab.~\ref{tab:benchmarks} the addition of our CMod module with global context guidance increases consistently the performance of the models on all the metrics \eg, a PSNR increase of $\approx\!2.0$dB. Note that because our CMod focuses on color reproduction (separately for the main model), this decreases the need for a complex reconstruction model. In other words, thanks to the CMod module, we do not require complex and deep neural networks to model the ISP transformation.

Our proposed ISP with basic building blocks and $100\times$ fewer parameters performs on par with the \emph{improved} state-of-the-art models. This shows the effectiveness of our CMod module in increasing the performance while decreasing the required overall model complexity. 

\begin{figure*}[!ht]
    \centering
    \setlength{\tabcolsep}{1pt}
    \begin{tabular}{c c c}
         \includegraphics[width=0.32\textwidth]{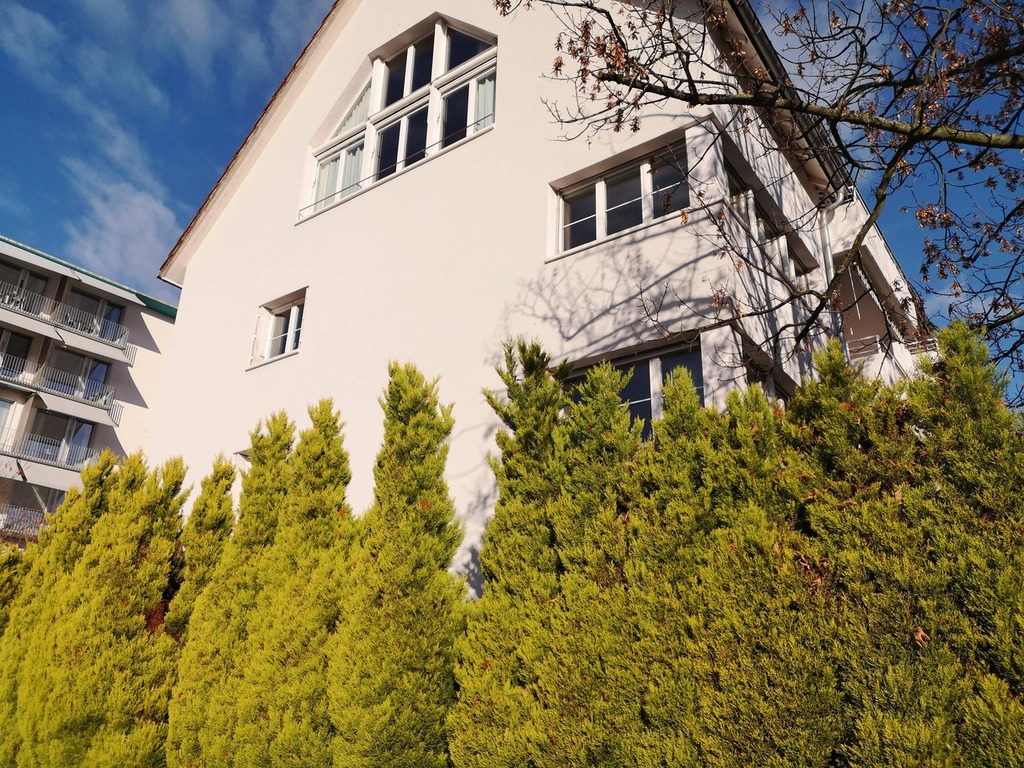} &
         \includegraphics[width=0.32\textwidth]{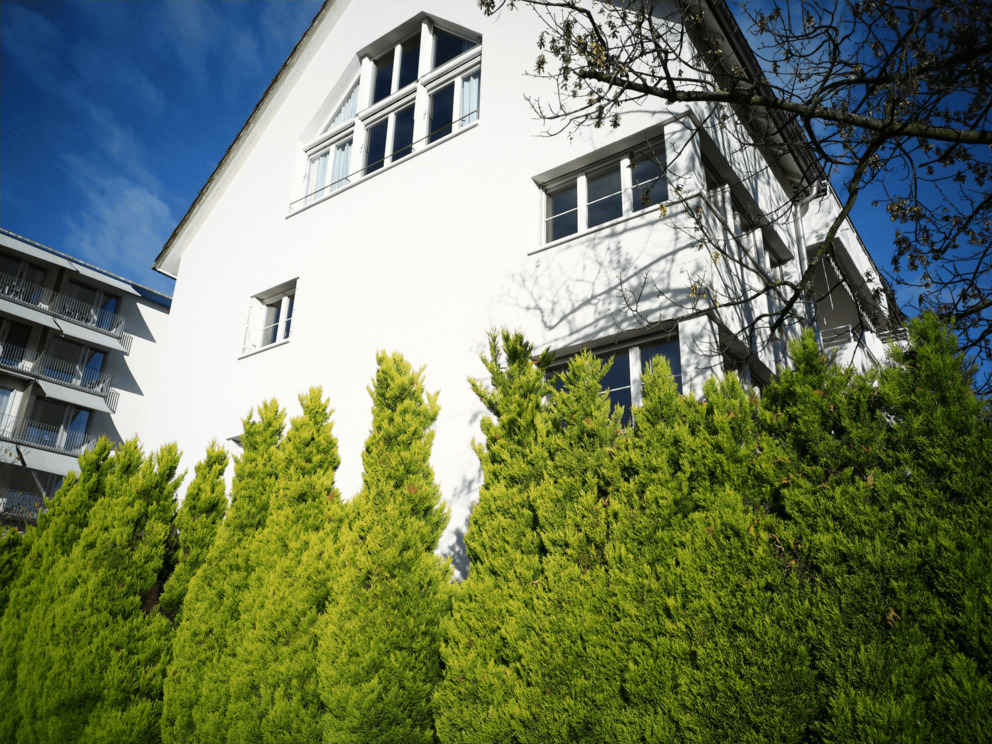} &
         \includegraphics[width=0.32\textwidth]{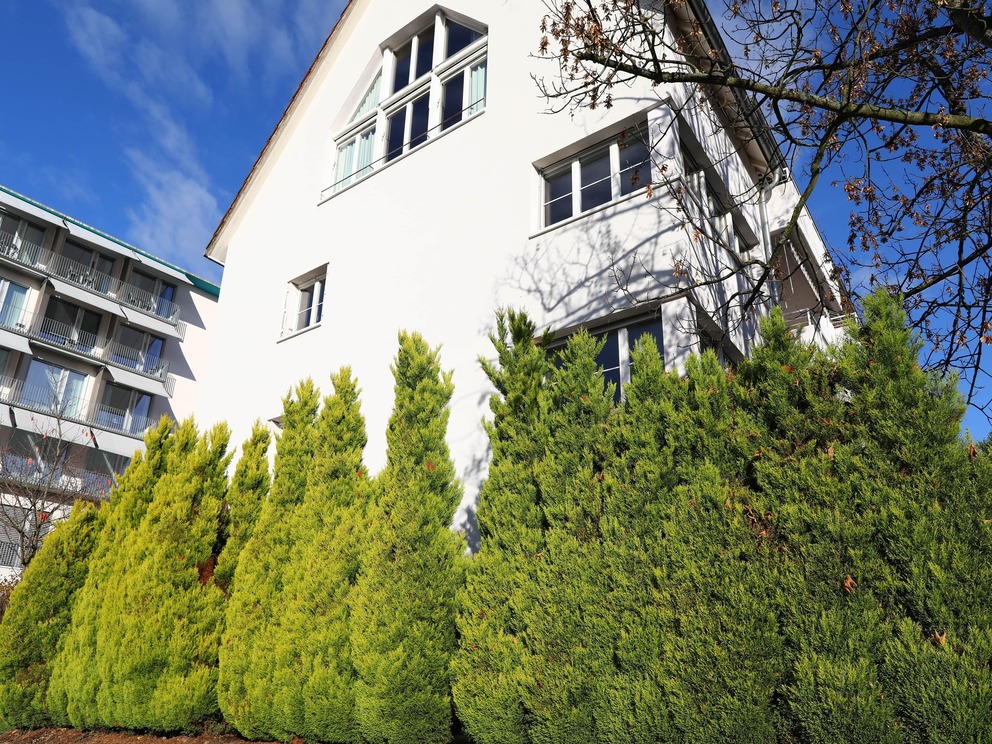} \\
         \includegraphics[width=0.32\textwidth]{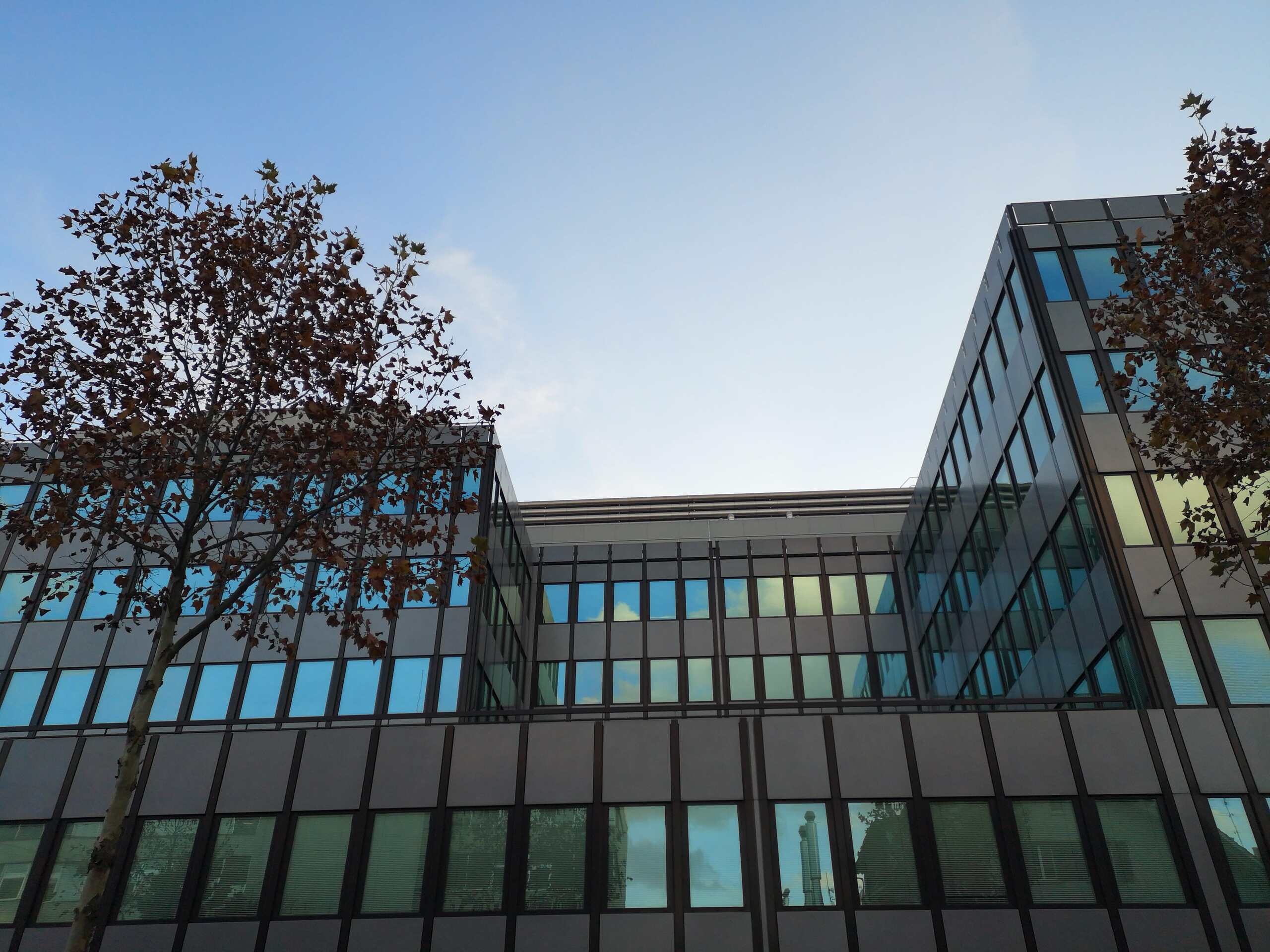} &
         \includegraphics[width=0.32\textwidth]{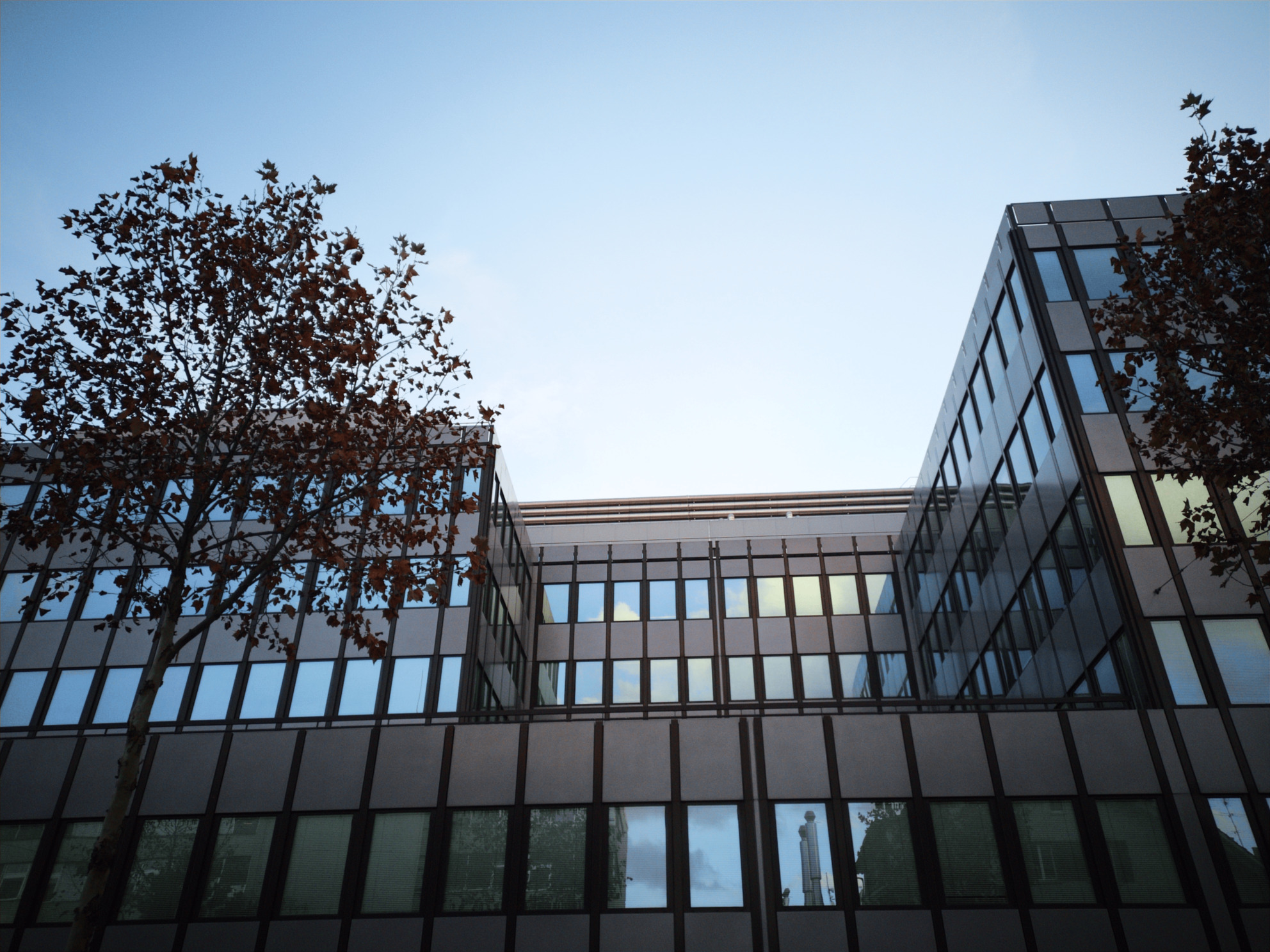} &
         \includegraphics[width=0.32\textwidth]{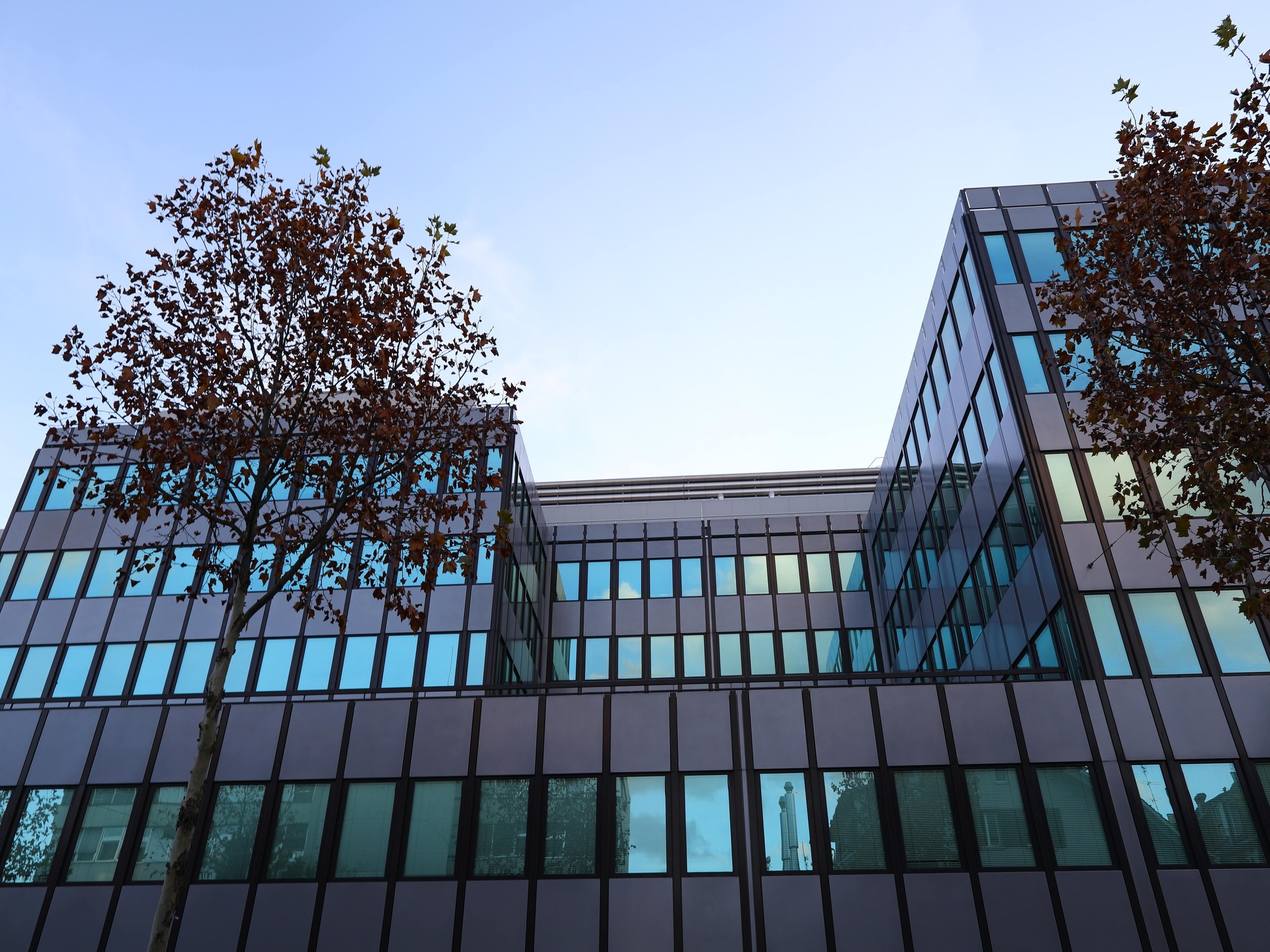} \\
         Huawei ISP & \ours \emph{(ours)} & Reference DSLR \\
    \end{tabular}
    \caption{Sample of full-resolution image processed using our \ours on the \emph{\textbf{ZRR Dataset}~\cite{Ignatov2020pynet} (Huawei P20)}. Our method processes the same RAW image as the smartphone-integrated ISP. We can get closer to the DSLR appearance.}
    \label{fig:fullimg}
\end{figure*}

\begin{figure*}[!ht]
    \centering
    \resizebox{.90\textwidth}{!}{
    \setlength{\tabcolsep}{1pt}
    \begin{tabular}{c c c}
         \includegraphics[width=0.33\textwidth]{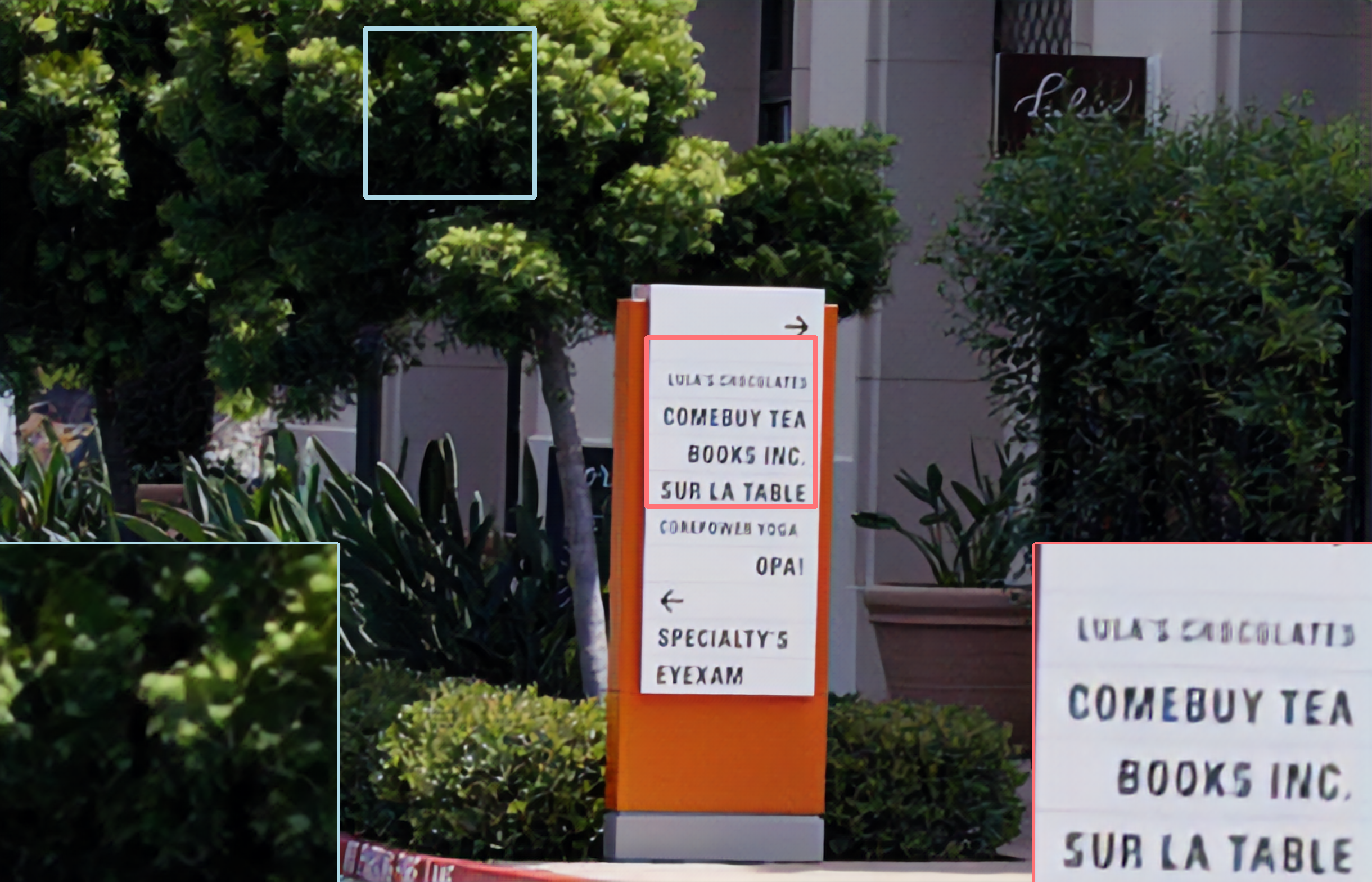} &
         \includegraphics[width=0.33\textwidth]{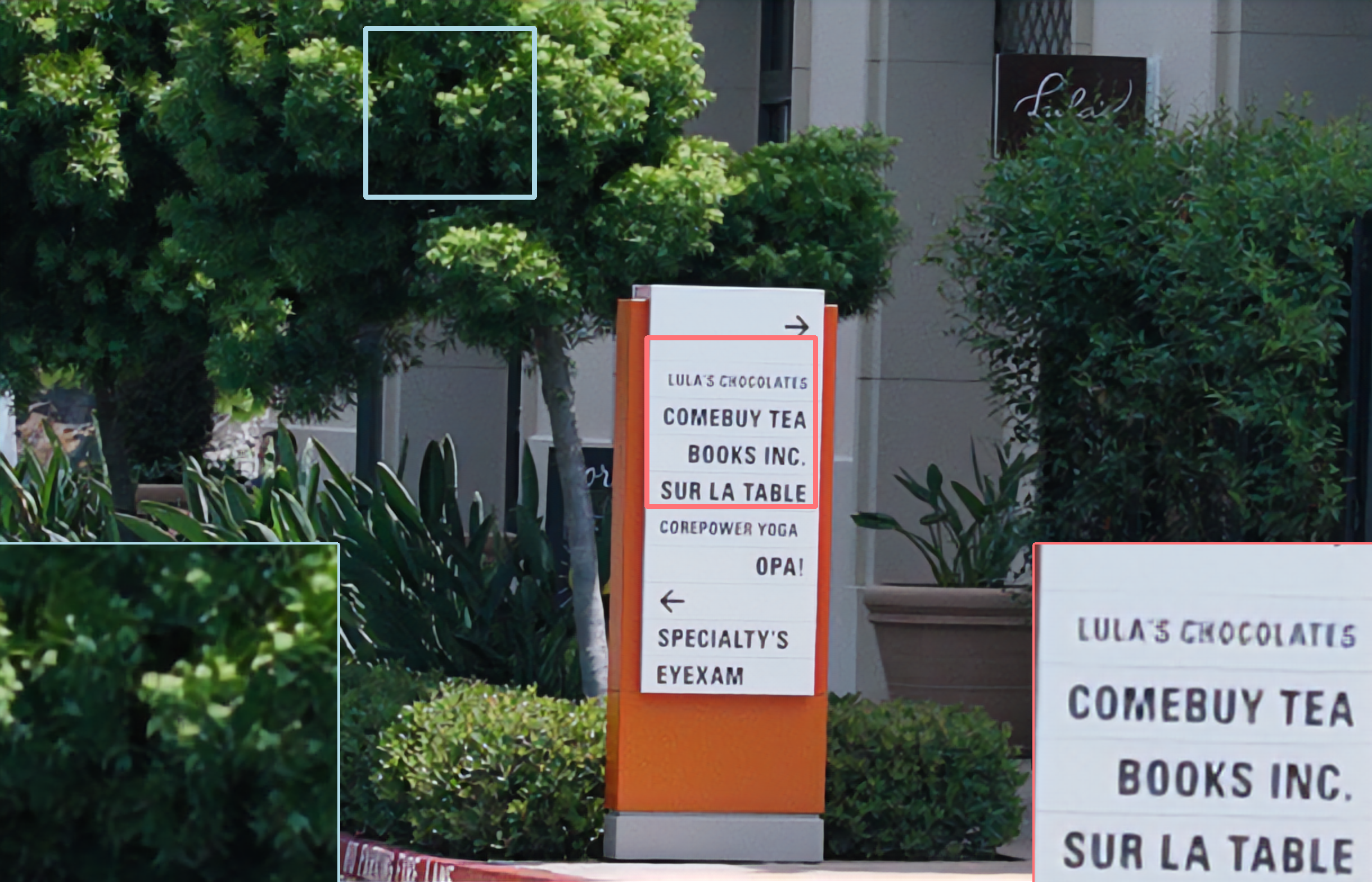} &
         \includegraphics[width=0.33\textwidth]{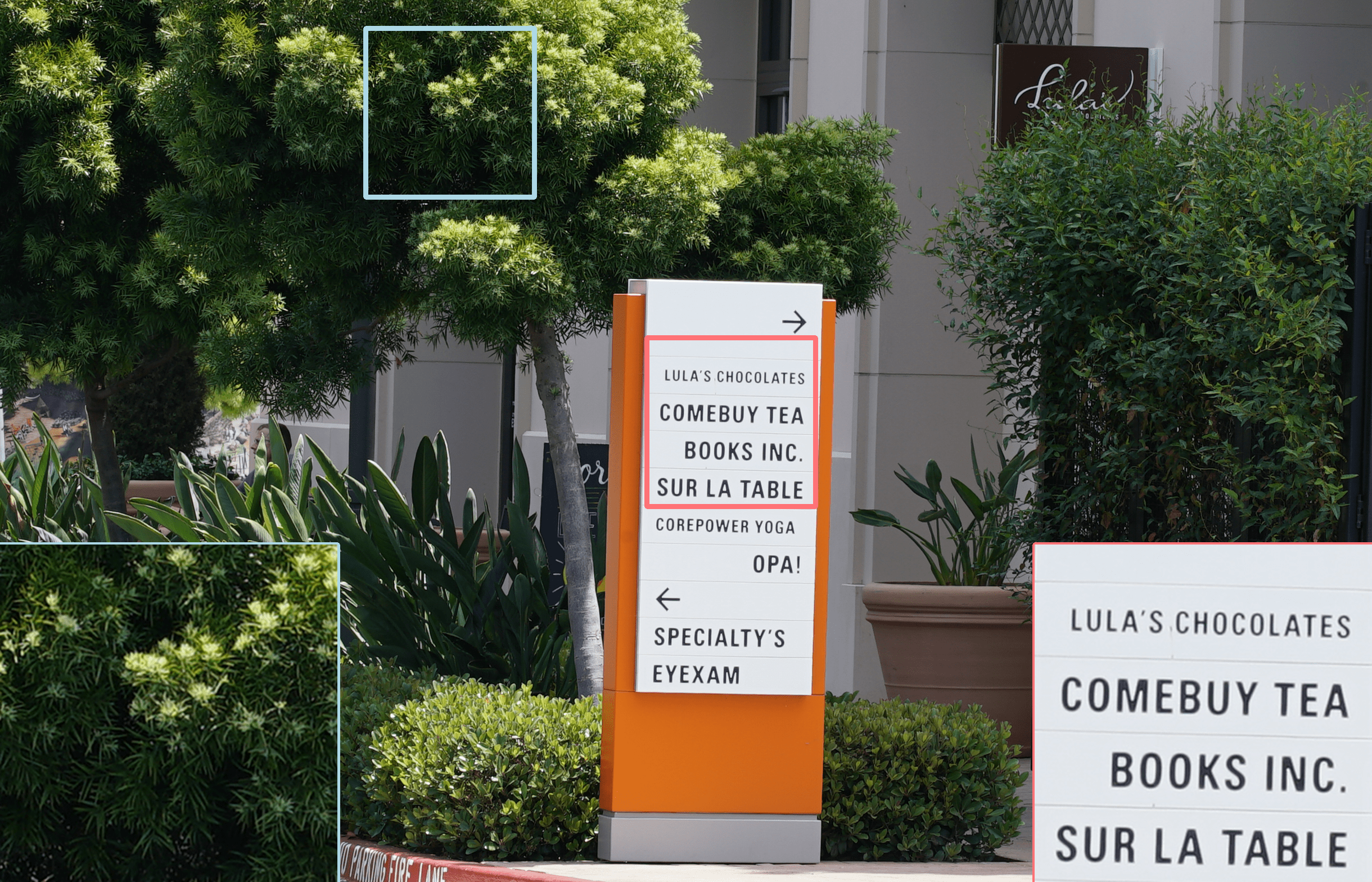} \\
         \includegraphics[width=0.33\textwidth]{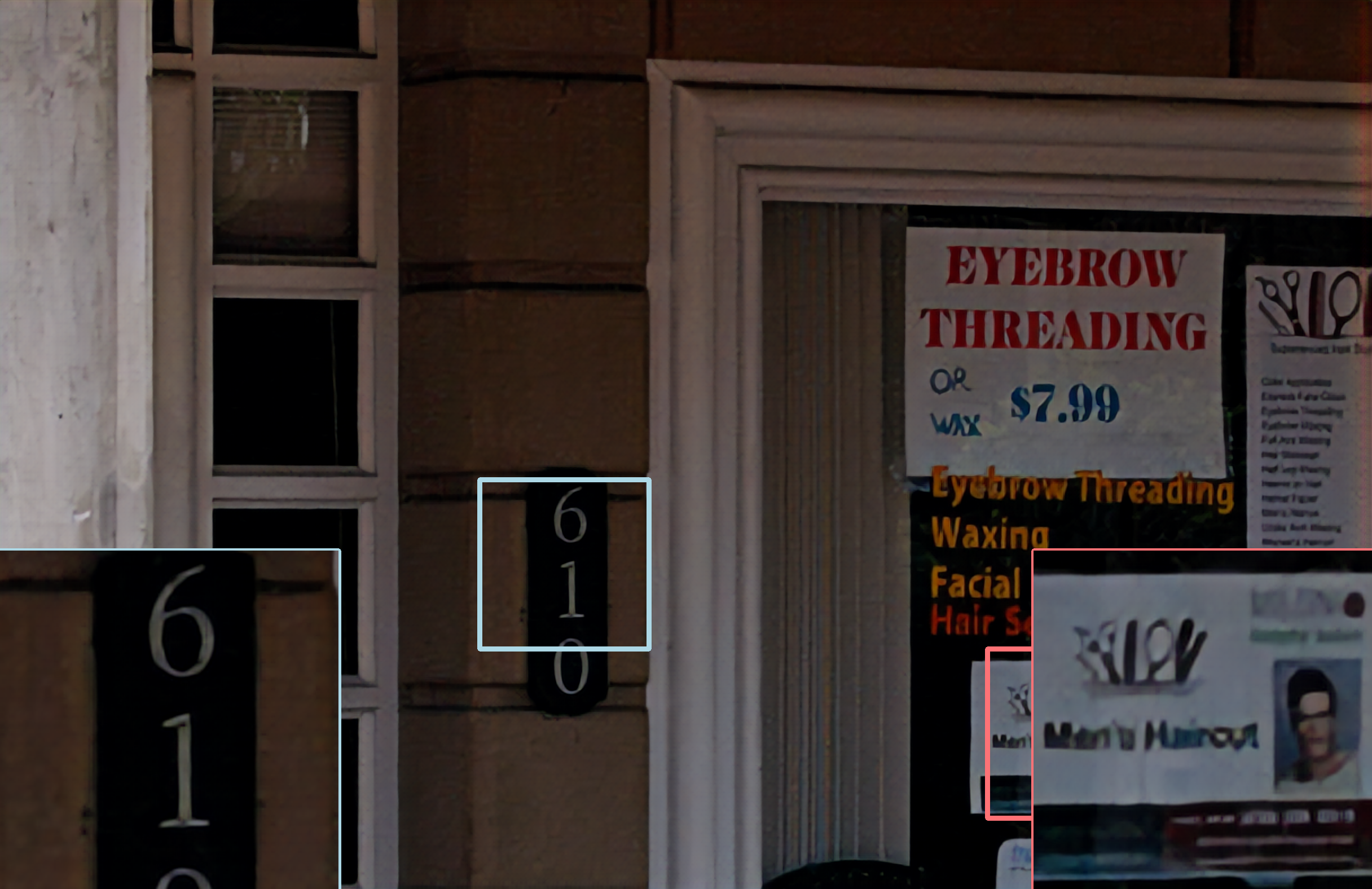} &
         \includegraphics[width=0.33\textwidth]{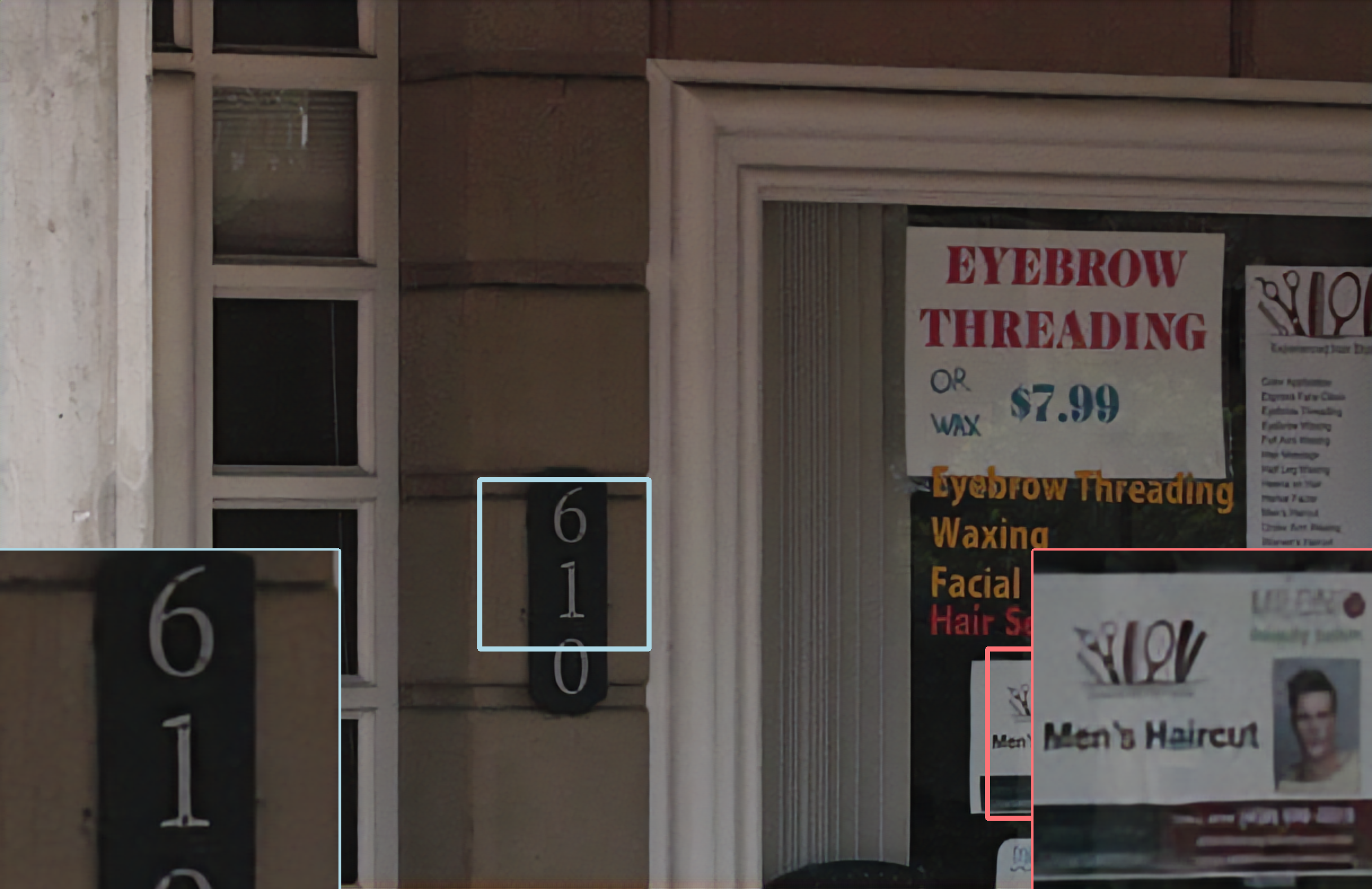} &
         \includegraphics[width=0.33\textwidth]{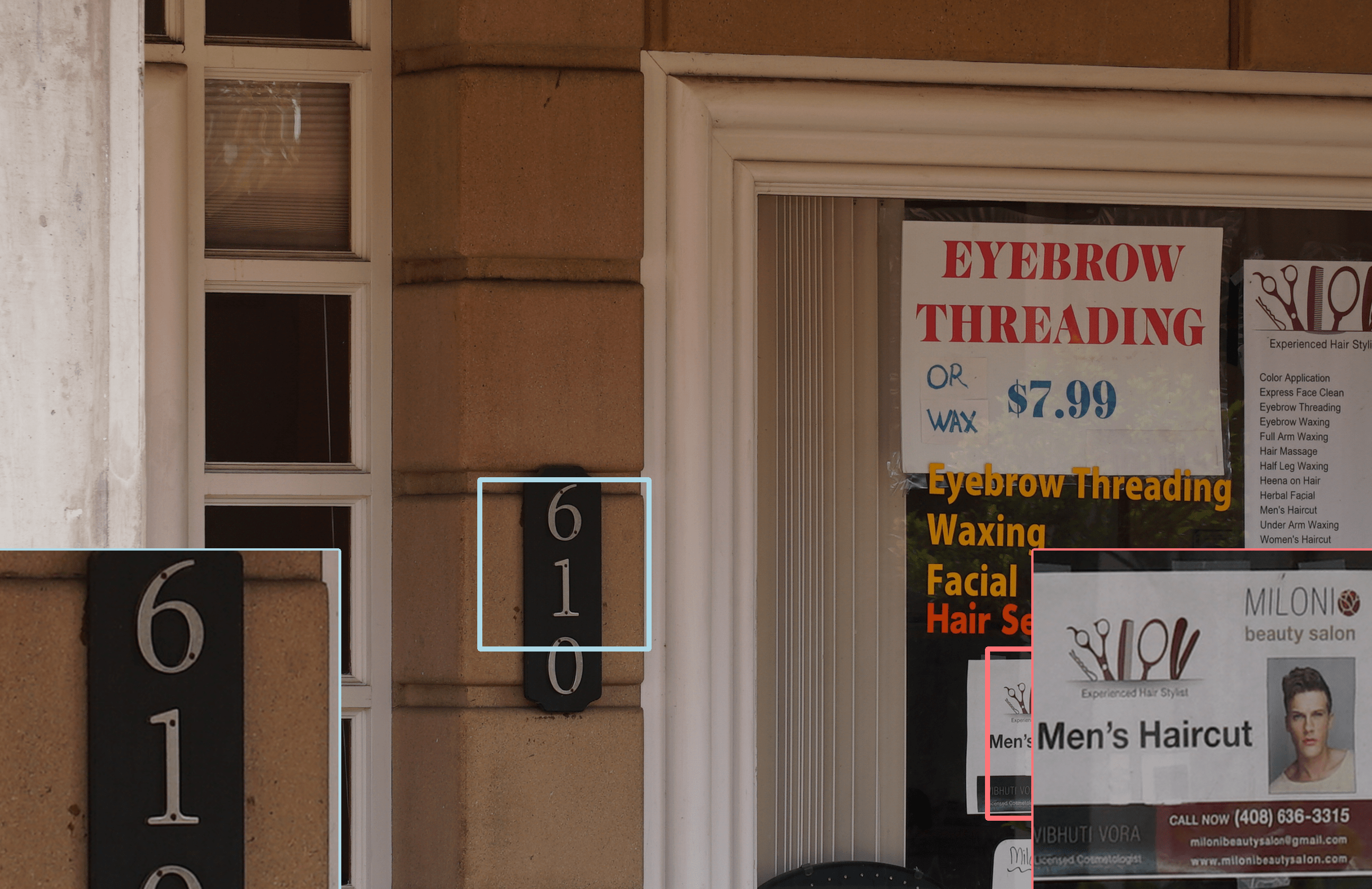} \\
         \includegraphics[width=0.33\textwidth]{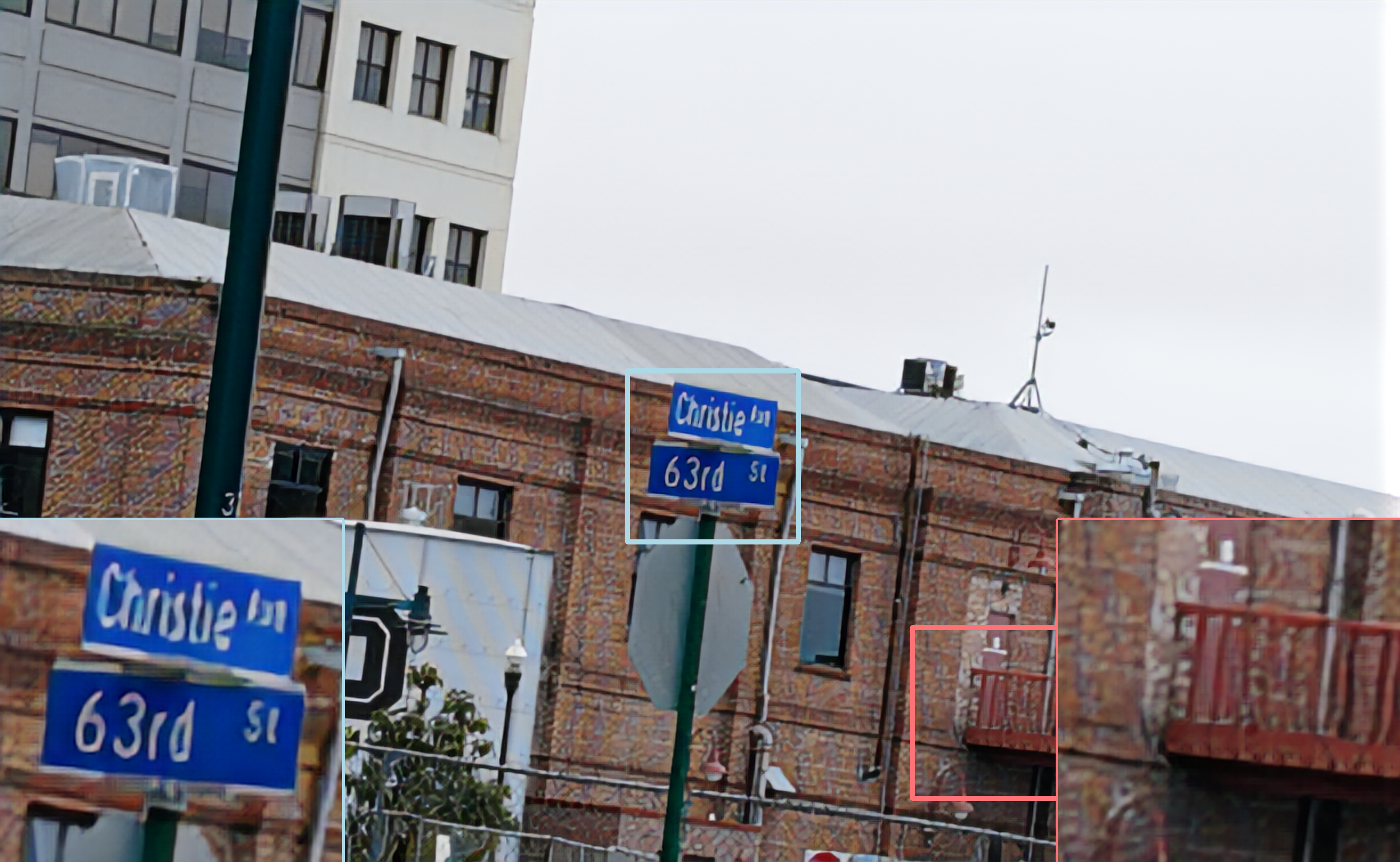} &
         \includegraphics[width=0.33\textwidth]{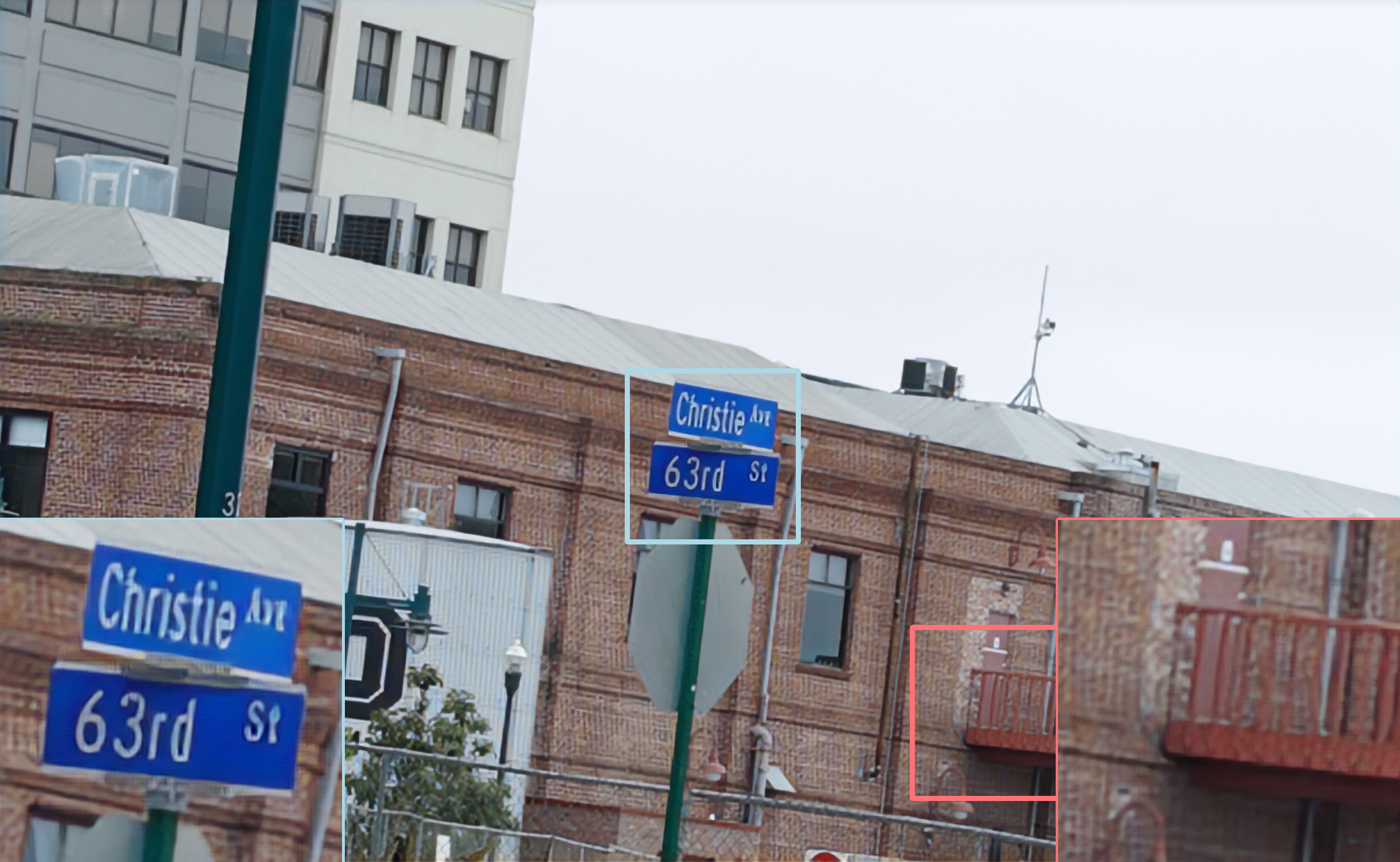} &
         \includegraphics[width=0.33\textwidth]{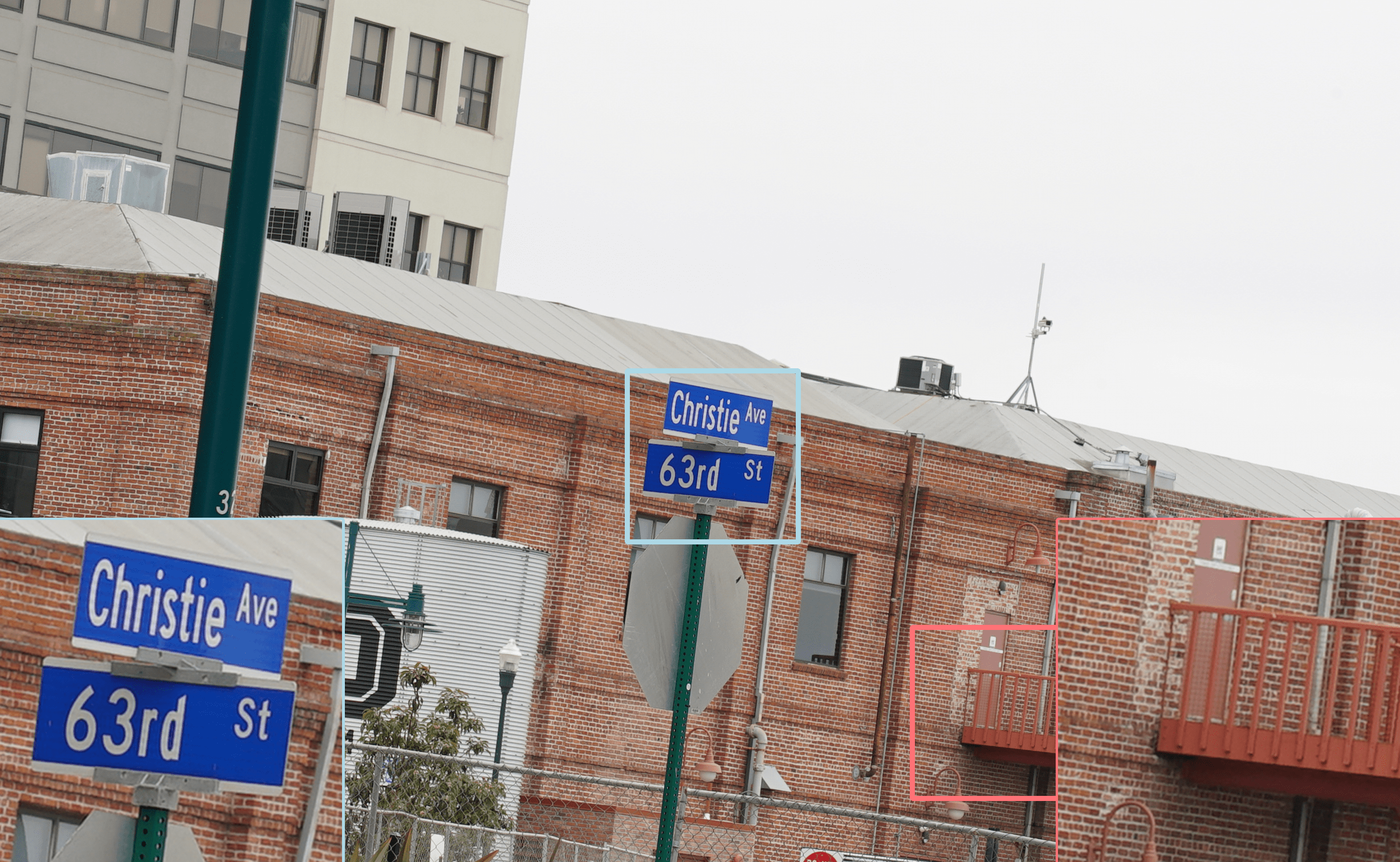} \\
        Zhang \etal~\cite{zhang2021learning} &  Ours & Reference DSLR \\
    \end{tabular}
    }
    \caption{Qualitative comparison of RAW Super Resolution Network using the \emph{\textbf{SR-RAW Dataset}~\cite{zhang2019zoom}}.}
    \label{fig:sraw-results}
\end{figure*}

\begin{figure*}[!ht]
    \centering
    \resizebox{.9\textwidth}{!}{
    \setlength{\tabcolsep}{1pt}
    \begin{tabular}{c c c c c}
         \includegraphics[width=0.19\textwidth]{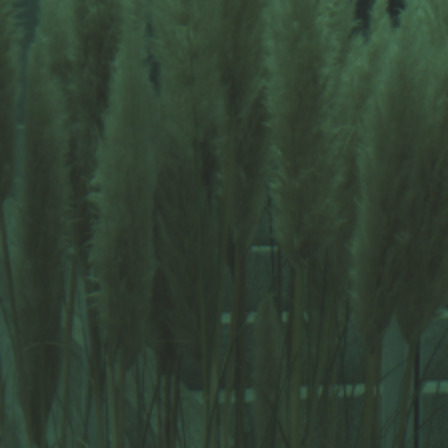} &
         \includegraphics[width=0.19\textwidth]{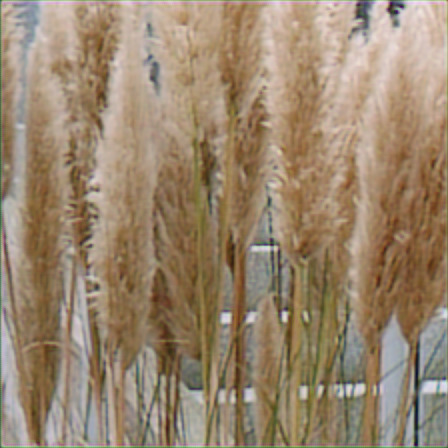} &
         \includegraphics[width=0.19\textwidth]{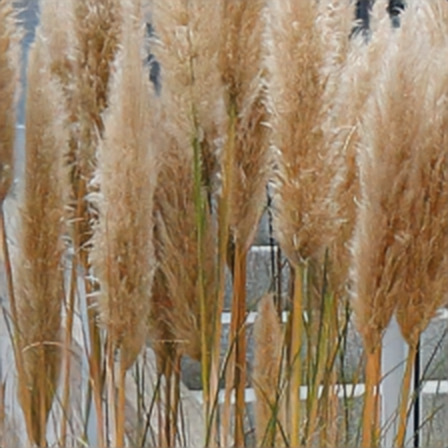} &
         \includegraphics[width=0.19\textwidth]{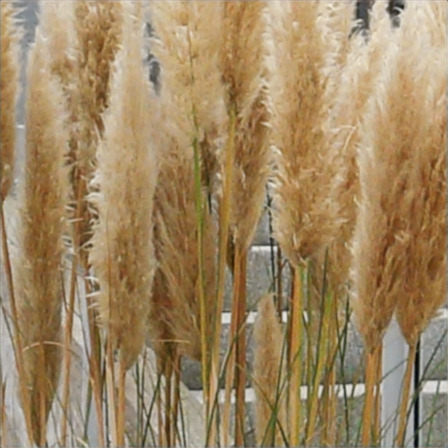} &
         \includegraphics[width=0.19\textwidth]{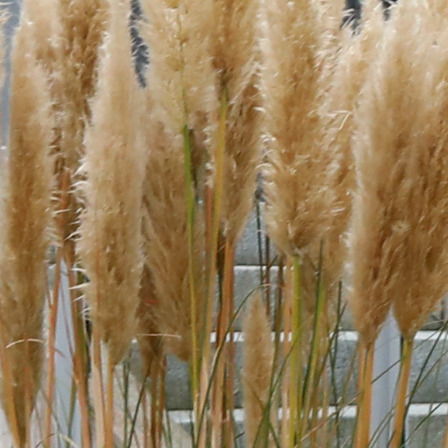} \\
         \includegraphics[width=0.19\textwidth]{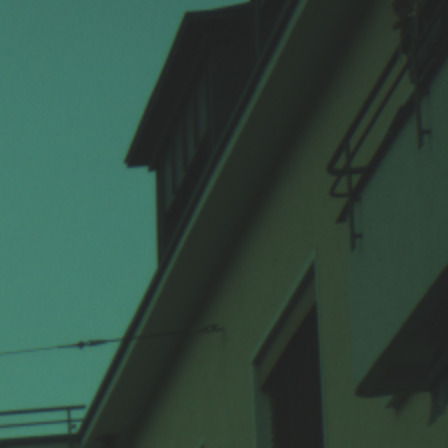} &
         \includegraphics[width=0.19\textwidth]{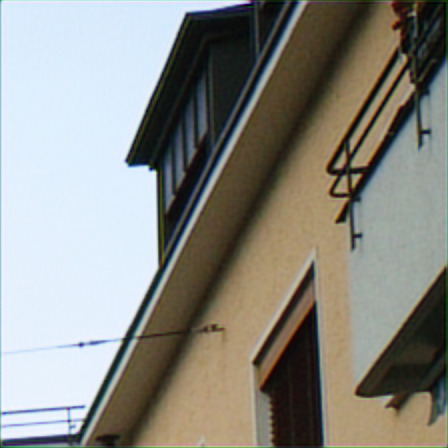} &
         \includegraphics[width=0.19\textwidth]{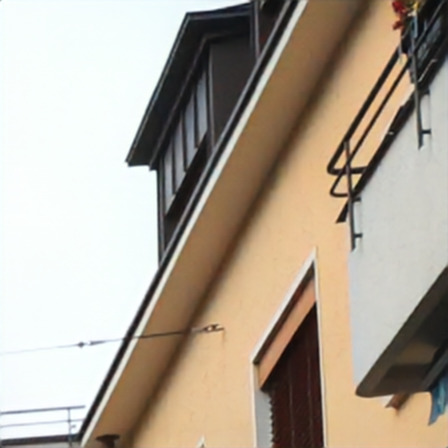} &
         \includegraphics[width=0.19\textwidth]{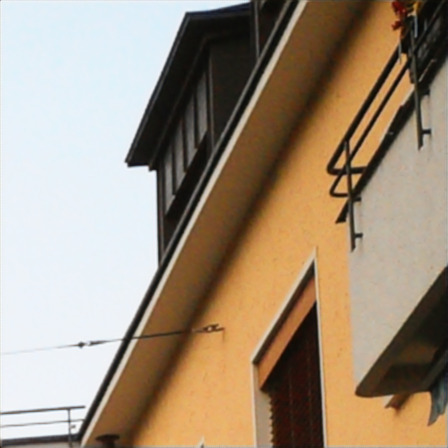} &
         \includegraphics[width=0.19\textwidth]{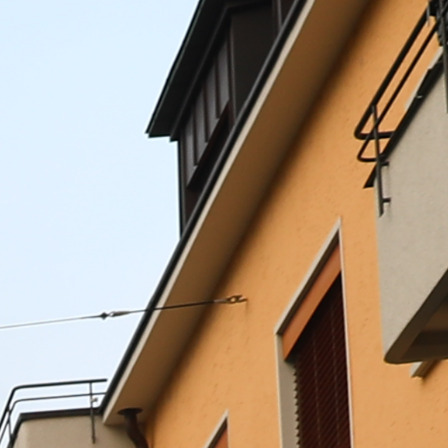} \\
         RAW Phone & MicroISP~\cite{ignatov2022microisp} & LiteISP~\cite{zhang2021learning} & \ours~\emph{(ours)} & Reference DSLR \\
    \end{tabular}
    }
    \caption{Qualitative comparison of neural ISPs using the \emph{\textbf{ZRR Dataset}~\cite{Ignatov2020pynet} (Huawei P20)}. Our \ours achieves the best color mapping. Best viewed in electronic version. Note that images are slightly compressed.}
    \label{fig:zrr-results}
\end{figure*}

\begin{figure*}[!ht]
    \centering
    \resizebox{.9\textwidth}{!}{
    \setlength{\tabcolsep}{1pt}
    \begin{tabular}{c c c c c}
         \includegraphics[width=0.19\textwidth]{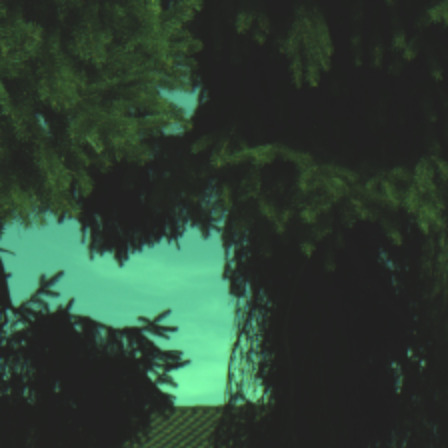} &
         \includegraphics[width=0.19\textwidth]{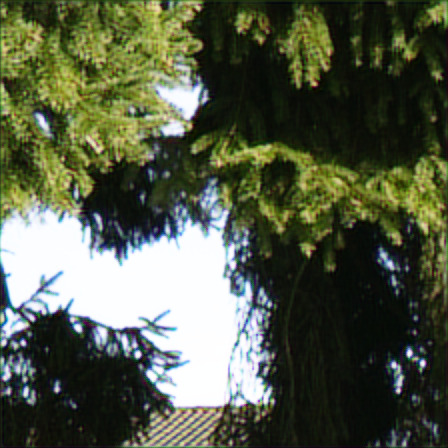} &
         \includegraphics[width=0.19\textwidth]{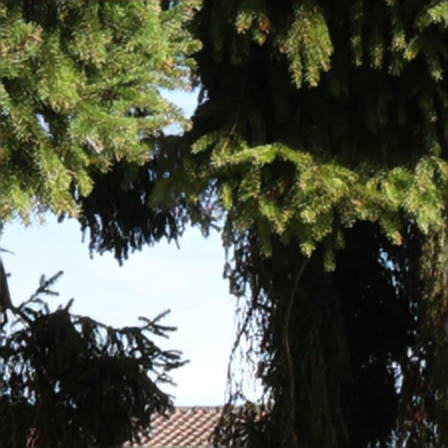} &
         \includegraphics[width=0.19\textwidth]{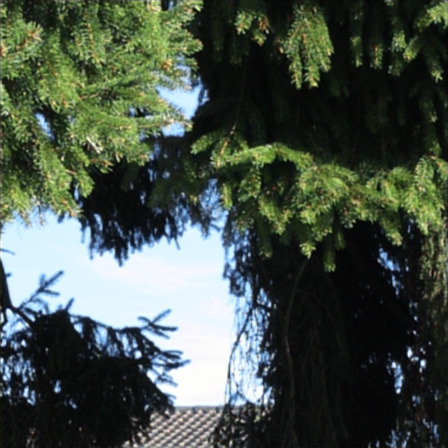} &
         \includegraphics[width=0.19\textwidth]{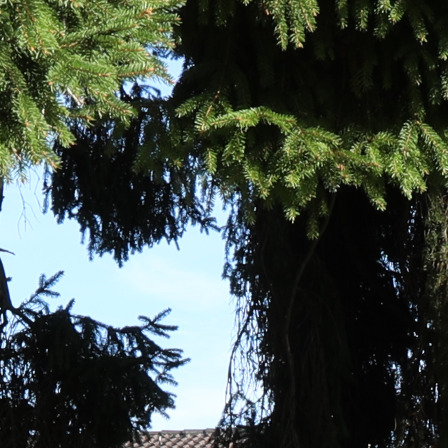} \\
         \includegraphics[width=0.19\textwidth]{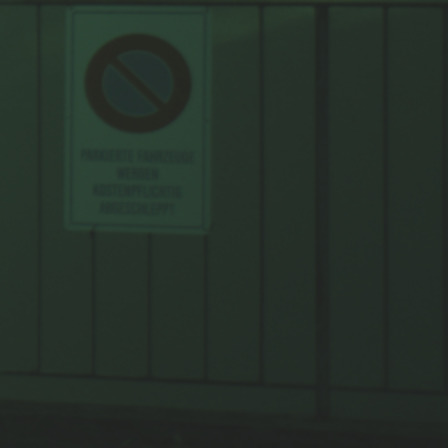} &
         \includegraphics[width=0.19\textwidth]{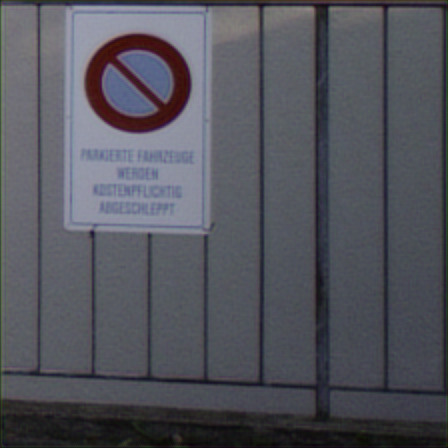} &
         \includegraphics[width=0.19\textwidth]{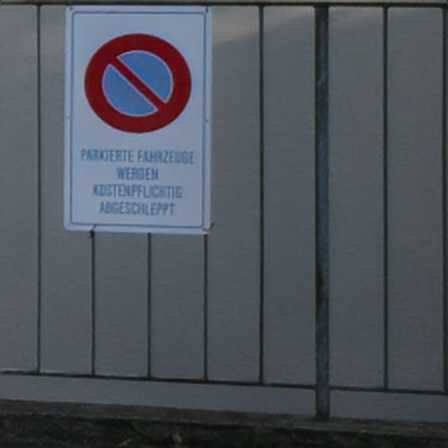} &
         \includegraphics[width=0.19\textwidth]{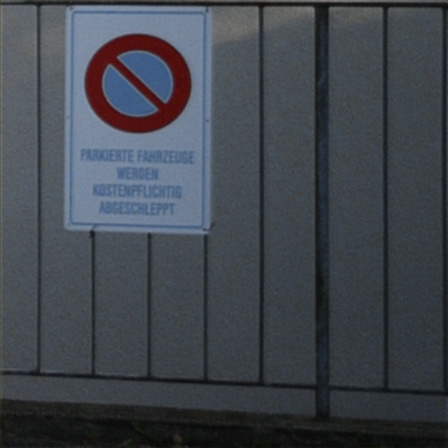} &
         \includegraphics[width=0.19\textwidth]{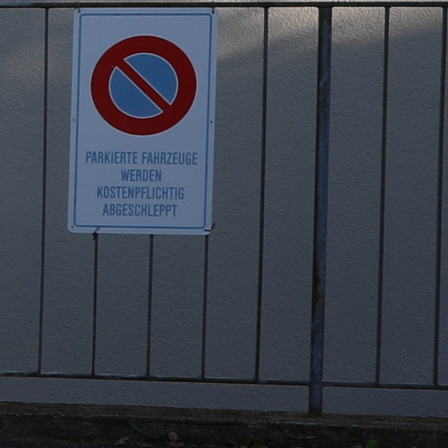} \\
         RAW Phone & MicroISP~\cite{ignatov2022microisp} & LiteISP~\cite{zhang2021learning} & \ours~\emph{(ours)} & Reference DSLR \\
    \end{tabular}
    }
    \caption{Qualitative comparison of neural ISPs using the \emph{\textbf{ ISPIW Dataset}~\cite{shekhar2022transform} (Huawei Mate 30 Pro)}.}
    \label{fig:ispw-results}
\end{figure*}

\vspace{2mm}
\textbf{Qualitative Results}: We provide exhaustive qualitative results in Figures~\ref{fig:fullimg},~\ref{fig:sraw-results},~\ref{fig:zrr-results}, and \ref{fig:ispw-results}. In Fig.~\ref{fig:fullimg} we show full-resolution processed RGB images. Our \ours model can reproduce the DSLR colors better than the original phone's ISP both processing the same RAW image.

As we notice from Figures~\ref{fig:zrr-results} and \ref{fig:ispw-results}, our model has a much better color rendering with more natural colors, better tone mapping with less over-exposure, and improved dynamic range. We can also see how our model produces sharper images, which shows the effectiveness of having two separate processing stages for global tasks and local tasks.

\subsection{Limitations}

Despite the promising results, we acknowledge certain limitations. First, due to the small training dataset, the models struggle to generalize in complex scenes \eg high dynamic range, high contrast, and detailed textures. Second, we do not employ complex attention mechanisms to capture global interactions (\eg Transformers) across the image, yet these techniques could help to achieve better performance. Third, the model struggles with spatially-varying effects such as lens shading. We believe that integrating positional encoding will help the model to understand better the global scene.

Finally, deploying the models on mobile devices usually requires a conversion to \textsc{TFLite} or \textsc{ONNX}, however, some operations might not be convertible (or supported by the target device), and the models are prone to losing performance.

\section{Discussion and concluding remarks}
We propose a simple ISP baseline model that captures the global context of the image to guide the image processing steps. An explicit separation of global and local transformations allows us to derive a compact and powerful model. Our model, \ours, achieves state-of-the-art results on different benchmarks using diverse and real smartphone images. Moreover, our global guidance module (CMod) can be integrated into any neural ISP to enhance it. Therefore our work represents a new baseline and benchmark for learned ISPs.

\vspace{2mm}
\noindent\textbf{Future Work} We plan to release more models and datasests for learned and controllable ISP.

\vspace{2mm}
\noindent\textbf{Reproducibility statement} \emph{Our code and models will be open-source at \url{https://github.com/mv-lab/AISP}.}

\vspace{2mm}
\noindent\textbf{Acknowledgments} This work was partly supported by the The Humboldt Foundation (AvH).


\begin{figure*}[!ht]
     \centering
     \setlength{\tabcolsep}{1pt}
     \begin{tabular}{c c c c}
          \includegraphics[width=0.24\textwidth]{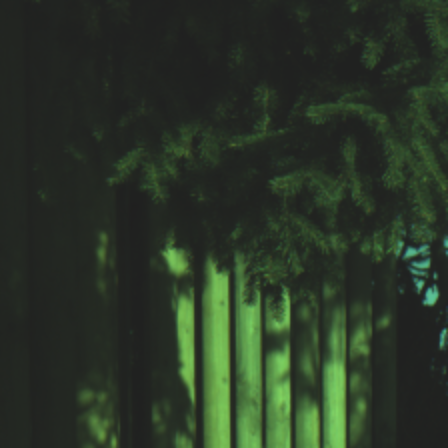} &
          \includegraphics[width=0.24\textwidth]{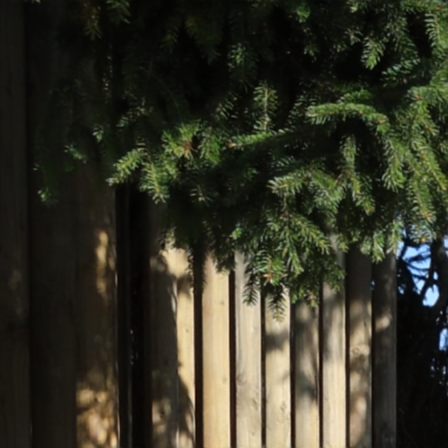} &
          \includegraphics[width=0.24\textwidth]{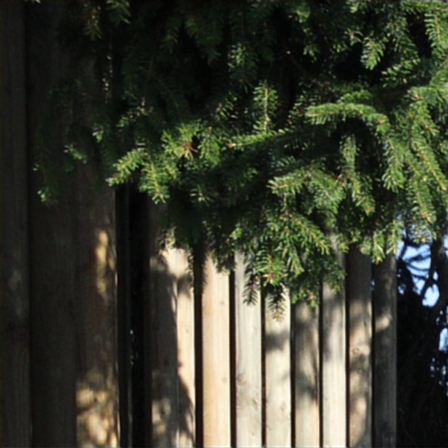} &
          \includegraphics[width=0.24\textwidth]{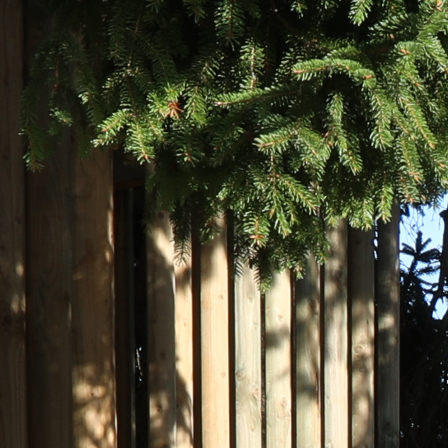} \\
          RAW Phone & LiteISP~$\dagger$ & \ours & Reference DSLR \\
     \end{tabular}
     \caption{Visual ablation study of \textbf{our methods}. LiteISP~$\dagger$~\cite{zhang2021learning} was trained using our global context attention. Our neural ISP, \ours is $20\times$ smaller than other methods. Sample from the \emph{\textbf{ISPIW Dataset}~\cite{shekhar2022transform}}.}
     \label{fig:ours}
\end{figure*}




{
\small
\bibliographystyle{IEEEbib}
\bibliography{main}
}
\end{document}